\documentclass[journal]{IEEEtran}
\usepackage{ifpdf}

\ifCLASSOPTIONcompsoc
 \usepackage[caption=false,font=normalsize,labelfont=sf,textfont=sf]{subfig}
\else
 \usepackage[caption=false,font=footnotesize]{subfig}
\fi

\usepackage{array}
\usepackage{csquotes}

\usepackage[pdftex]{graphicx}
\usepackage{epstopdf}
\graphicspath{{../figs/}}
\DeclareGraphicsExtensions{.pdf,.jpeg,.png,.eps,.gif}

\usepackage{amsthm}
\usepackage{amsfonts}
\usepackage{amsmath} 
\usepackage{amssymb}  
\usepackage[italicdiff]{physics}
\usepackage[per-mode=symbol]{siunitx}
\theoremstyle{plain}

\theoremstyle{definition}
\newtheorem{defn}{Definition}

\theoremstyle{remark}
\newtheorem{rem}{Remark}

\usepackage{cite}

\usepackage{algorithm}
\usepackage{algorithmicx}
\usepackage{algpseudocode}

\usepackage{url}
\usepackage{xcolor}
\usepackage[]{hyperref}
\hypersetup{
	pdfinfo={
		pdfproducer={},
		Title={},
		Subject={},
		Author={},
	}
}

\usepackage{cleveref}

\usepackage[mathlines,columnwise,switch]{lineno}

\begin{document}


\title{From the logistic-sigmoid to \textbf{n}logistic-sigmoid: modelling the COVID-19 pandemic growth}
%
%
\author{Oluwasegun A. Somefun$^{\star\,1}$ Kayode F. Akingbade$^{2}$ and Folasade M. Dahunsi$^{1}$
\thanks{*This work was not supported by any organization}
\thanks{$^{\star\,1}$Oluwasegun Somefun {\tt\footnotesize oasomefun@futa.edu.ng}, and Folasade Dahunsi  {\tt\footnotesize  fmdahunsi@futa.edu.ng} are with the Department of Computer Engineering, Federal University of Technology Akure, PMB 704 Ondo, Nigeria.
}%
\thanks{$^{2}$Kayode Akingbade {\tt\footnotesize kfakingbade@futa.edu.ng} is with the Department of Electrical and Electronics Engineering, Federal University of Technology Akure, PMB 704 Ondo, Nigeria.
}%
}

\maketitle



\begin{abstract}
Real-world growth processes, such as epidemic growth, are inherently noisy, uncertain and often involve multiple growth phases. The logistic-sigmoid function has been suggested and applied in the domain of modelling such growth processes. However, existing definitions are limiting, as they do not consider growth as restricted in two-dimension. Additionally, as the number of growth phases increase, the modelling and estimation of logistic parameters becomes more cumbersome, requiring more complex tools and analysis. To remedy this, we introduce the \textbf{n}logistic-sigmoid function as a compact, unified modern definition of logistic growth for modelling such real-world growth phenomena. Also, we introduce two characteristic metrics of the logistic-sigmoid curve that can give more robust projections on the state of the growth process in each dimension.  Specifically, we apply this function to modelling the daily World Health Organization published COVID-19 time-series data of infection and death cases of the world and countries of the world to date. Our results demonstrate statistically significant goodness of fit greater than or equal  to 99\% for affected countries of the world exhibiting patterns of either single or multiple stages of the ongoing COVID-19 outbreak, such as the USA. Consequently, this modern logistic definition and its metrics, as a machine learning tool, can help to provide clearer and more robust monitoring and quantification of the ongoing pandemic growth process.
\end{abstract}

\begin{IEEEkeywords}
Logistic function, Sigmoid; Nonlinear Functions; Neural Networks; Growth; Process; Dynamics; COVID-19; Non-linear Regression; Time-series
\end{IEEEkeywords}

\section{Introduction}\label{sec_intro}

We introduce a modern approach to modelling real-world growth (or decay) processes by extending the classic logistic-sigmoid function definition -- the \textbf{n}logistic-sigmoid function (\textsc{nlsig}). Motivated by observing the multiple sigmoidal paths of ants, and that the growth (or motion) of a phenomenon  with respect to its input (such as time) is always finite in both input--output dimensions, we propose the \textsc{nlsig}, a compact and unified logistic-sigmoid function (\textsc{lsig}) definition for sigmoidal curvatures with multiple peak inflection points. Such a function in the general multiple-input multiple-output sense, can be viewed as equivalent to a feed-forward neural network, and has the notable advantage of improving the modelling power of the logistic-sigmoid function. In addition, we introduce two characteristic metrics of the \textsc{nlsig} curve that can give more robust real-time projective measures on the geometric (exponential) state of a growth process. We demonstrate how the basic \textsc{nlsig} neural pipeline can be applied to modelling and monitoring the COVID-19 pandemic, which is a current real-world (natural) growth phenomena. COVID-19 data-sets (infections and death cases) obtained from the World Health Organization were used. The \textsc{nlsig} model demonstrated up-to 99\% statistical significant goodness-of-fit in our experiments. The \textsc{nlsig} code for this paper is available at the repository \cite{somefunSomefunAgbaNLSIGCOVID19Lab2020}. Consequently, as the logistic-sigmoid definition is unified through the \textsc{nlsig} with enhanced modelling power, we anticipate that it will find immense benefit in the many scientific areas were predictive growth-modelling and regression analysis of real-world phenomena (single or multiple outcomes) are of interest.

\subsection{Background}\label{sec_backgnd}

\textbf{Classic Logistic-Sigmoid}. As opposed to an infinite (or unrestricted) exponential (or logarithmic) growth function, the logistic growth function is a finite (or restricted) logarithmic growth function, hence the name `logistic' \cite{shulmanMathAliveUsingOriginal1998,jefferyRestrictedLogarithmicGrowth2016}. There are many descriptions that can be given to the logistic-sigmoid function. One is that: it is a step-function that features a convex-concave ($\mathcal{S}$) or concave-convex (mirrored $\mathcal{S}$) shape with a derivative function having a bell (or normal)-like distribution \cite{udellMaximizingSumSigmoids2013}. Such a bell-shape becomes triangular when data is coarse. Classically, it is defined or modelled as having a output domain of finite minimum and maximum bounds \(\left[y_{\min},y_{\max}\right]\) on an input co-domain (support) of infinite bounds \(\left[-\infty,\infty\right]\) respectively \cite{cybenkoApproximationSuperpositionsSigmoidal1989,lipovetsky_double_2010}.
\begin{IEEEeqnarray}{c}
	y = \mathbf{\sigma}\left(\mathrm{x}\right)=y_{\min}+\frac{y_{\max}-y_{\min}}{\left(1 + e^{\pm\alpha\left( \mathrm{x}-\delta\right)}\right)}\label{lsig}
\end{IEEEeqnarray}
Logistic-sigmoids have found numerous useful scientific and engineering applications to real-world data. This is evident  in probability-theory and statistics such as, for example as activation functions in artificial neural networks, logistic regression models, resource allocation models, maximum utility models, population-growth models, plant-growth models, autocatalysis reaction models, signal-processing models, power-transistor models, fermi-dirac distribution models, tumour and infection growth models, in fields such as: engineering and computer science, economics, bioinformatics, population biology, crop-science, chemistry, semi-conductor physics, epidemiology, and many others for modelling certain real-world growth processes \cite{szeEfficientProcessingDeep2017,barkerLogisticFunctionComplementary2020,christopoulosEfficientIdentificationInflection2016,yeoModellingTechniqueUtilizing2016,hemmatiFairEfficientBandwidth2016,xsRichardsModelRevisited2012,udellMaximizingSumSigmoids2013,shaoNonlinearTrackingDifferentiator20140901,yinFlexibleSigmoidFunction2003,leeEstimationCOVID19Spread2020,lipovetsky_double_2010}. Notably, the logistic-sigmoid function is characterized by a point of maximum growth rate (peak inflection-point). Mathematically, this is described as the point where the second-derivative or acceleration changes sign, that is a value of zero. Therefore a necessary condition that a point $x_t$ on the support (dependent variable) interval is an inflection-point is that $  \sigma^{\prime\prime}\left( x_{t+1}  \right) $ and $  \sigma^{\prime\prime}\left( x_{t-1}  \right) $ have opposite signs.
\begin{IEEEeqnarray}{c}
	y = \mathbf{\sigma}\left(\mathrm{x}\right)=y_{\min}+\frac{y_{\max}-y_{\min}}{\left(\varsigma + \gamma e^{\pm\alpha\left( \mathrm{x}-\delta\right)}\right)^{\beta^{-1}}}\label{rlsig}
\end{IEEEeqnarray}

Interestingly, the logistic-sigmoid function (\ref{lsig}) has been modified and referred to by different names such as the generalised logistic model (\ref{rlsig}) or the generalised richard's model \cite{wuGeneralizedLogisticGrowth2020}. However, we note that this definition is over-parametrized by the presence of three parameters \(\varsigma, \gamma \) and \(\beta\), which violate the location integrity of \( \delta \) as the peak inflection-point and hence the maximum logistic growth fails to occur at the point $\mathrm{x}=\delta$, except when $\varsigma=\gamma=\beta=1$.  Primarily, the notion of the logistic-sigmoid being symmetric around $\delta$, as highlighted in \cite{yinFlexibleSigmoidFunction2003,loibelInferenceRichardsGrowth2006}, led to the richard's model to deal with both symmetric and asymmetric growth. Also, similar symmetry arguments are given in \cite{tabatabaiHyperbolasticGrowthModels2005}, where the authors propose hyperbolastic functions for generalizing sigmoidal growth. The authors fail to note that just like the gompertz function, error functions, trigonometric functions like the hyperbolic tangent, and so on lead to sigmoidal curvatures, they are all just over-complicated means in the `mathematical sense' of describing the standard logistic growth phenomenon.

Contrarily to these authors, we note that such notions on symmetry around the peak inflection-point becomes unfounded when the input co-domain bounds to the logistic-sigmoid function are enforced.
That is, when the realistic assumption that growth is always restricted, not only in the space dimension ($y$) of the phenomena but also in the time dimension ($\mathrm{x}$). Therefore, retaining its mathematical simplicity, the logistic-sigmoid function without the extra parametrizations of the richard's model or hyperbolastic models becomes by itself powerful enough to describe a wide range of natural growth-processes whether symmetric or not.

\textbf{Growth Processes}. The growth (flow or trend) dynamics in any direction (positive or negative) for many natural phenomena such as epidemic spreads, population growths, adoption of new ideas can be and have been approximately modelled by the logistic-sigmoid curve \cite{taylorForecastingScale2018}. The prevalence of the \textsc{lsig} is such that the cumulative and incident growth or flow distribution of many natural phenomena in the world follows a saturated non-linear growth well approximated by the sigmoidal and bell curve respectively \cite{yinFlexibleSigmoidFunction2003}. The scientific basis for this prevalence is, the constructal law origin of the \textsc{lsig}, given in \cite{bejanConstructalLawOrigin2011}. Many of such growth-processes can be viewed as complex input--output systems that involve multiple peak inflection phases with respect to time. Such dynamic systems can be modelled using a sum or superposition of logistic-sigmoids, an idea that can be traced back in the crudest sense to \cite{reedSummationLogisticCurves1927}. This idea is also seen, for instance, in the universal-approximation theorems perspective of artificial neural networks that state that: in principle, any continuous function can be approximated arbitrarily well, to any desired accuracy, by single-hidden layer networks formed by a suitably large finite additive combination of logistic-sigmoid functions between the input-output layer for pattern classification and non-linear time-series prediction tasks \cite{cybenkoApproximationSuperpositionsSigmoidal1989,hanInfluenceSigmoidFunction1995,udellMaximizingSumSigmoids2013}. Similarly, multistage logistic models have been discussed in \cite{barkerLogisticFunctionComplementary2020,batistaEstimationStateCorona2020,hsiehRealtimeForecastMultiphase2006,chowellNovelSubepidemicModeling2019,lipovetsky_double_2010} in application to modelling multiple waves of epidemics. In particular, the \textsc{lsig} with time-varying parameters is the core trend model in Facebook's Prophet model for time-series growth-forecasting at scale \cite{taylorForecastingScale2018} on big data.

However, two recurring limitations of the logistic definitions in these works exist, a trend that has continued since the first logistic-sigmoid function introduction.

First is that, the support interval (co-domain) of logistic-sigmoids is assumed to be infinite, this as we will discuss in the next section, violates the critical natural principle of finite growth. Second, estimation of the different hyper-parameters for the logistic-sigmoids that make the multiple logistic-sigmoid sum is done separately, instead of as a whole, single function. The effect of this, is that, as the number of logistic-sigmoids considered in the sum increases, regression analysis becomes more cumbersome and complicated as can be observed in these works \cite{leeEstimationCOVID19Spread2020, batistaEstimationStateCorona2020,hsiehRealtimeForecastMultiphase2006,wuGeneralizedLogisticGrowth2020,chowellNovelSubepidemicModeling2019,taylorForecastingScale2018}.

These limitations are efficiently overcome by the proposed \textsc{nlsig} function for the logistic-sigmoid curve with $n\ge1$ multiple peak inflections on a compactly bounded support interval with simplified computations of its derivatives and partial-derivatives at each layer, which make the \textsc{nlsig} compatible with gradient-descent optimization algorithms \cite{hanInfluenceSigmoidFunction1995,werbosBackpropagationTimeWhat1990}.

Admittedly, epidemiological models such as the SEIRD variants \cite{leeEstimationCOVID19Spread2020,okabeMathematicalModelEpidemics2020} are just another form of representing sigmoidal growth \cite{xsRichardsModelRevisited2012}. It has been noted in \cite{christopoulosNovelApproachEstimating2020} that the SEIRD-variant models yield largely exaggerated forecasts on the final cumulative growth number. This can also be said of the results of various application of logistic modelling as regards the current state of the COVID-19 pandemic which have largely resulted in erroneous identification of the epidemic's progress and its future projection \cite{matthewWhyModelingSpread2020, batistaEstimationStateCorona2020, wuGeneralizedLogisticGrowth2020}, hence leading policymakers astray. Compared to the basic-reproduction number metric that has been used in such epidemiological models \cite{xsRichardsModelRevisited2012} and engaging in false prophecy or predictions of an ongoing growth phenomena, whose source is both uncertain and complex to be encoded in current mathematical models, on the contrary, for the \textsc{nlsig} curve we introduce two reliable metrics for robust projective measurements of the modelled time-evolution of a growth-process in an area or locale of interest. Also, as opposed to most of the works, we use, as noted in \cite{christopoulosEfficientIdentificationInflection2016} the cumulative growth data, instead of the daily growth-rate data.

\section{The \textbf{n}logistic-sigmoid}\label{sec_thenlsig}

Exponential growth at its core implies an output function $y\to\infty$ as the input $\mathrm{x}\to\infty$. Interestingly, many real-world situations do not have infinite bounds or intervals, there is always a limit (or restriction) to everything. Approximation of this reality, led to the \textsc{lsig} growth model, in that, every phenomena is finite, even when it seems infinite. There is always a clear beginning and end, even if unknown. However, as can be observed in physical systems, the concept of $\mathrm{x}$, which we can call the time or input co-domain interval of a growth phenomenon, being finite is one that has been largely overlooked in the definitions of the \textsc{lsig} in the literature.

More generally, in this paper, we describe the \textsc{lsig} distribution as a map with: a single peak inflection-point which makes its derivative a bell-shape distribution; and two valley inflection-points which defines the minimum and maximum locations of its co-domain. The valley inflection-points indicate that the phenomena being modelled intrinsically has imposed limits
on its growth in a two-dimensional (2D) direction, which we can call output--input or space--time. Extending this description to $n$ peak inflection-points gives a unified modern definition of the \textsc{lsig}, which we term the \textbf{n}logistic-sigmoid (\textsc{nlsig}) function.

The logistic growth is, therefore, essentially a story of inflection-points on the $\mathrm{x}$ support interval. This implies that there are two possible ways to model such curves. One is by assuming that generally growth occurs on arbitrarily partitioned interval/sub-intervals or automating by assuming that growth occurs on an equally partitioned interval/sub-intervals. Since for real-world growth processes, events do not necessarily happen at equal intervals, the general definition, as will be presented, is a more powerful tool.

Specifically, using this notion of partitioning both the output and input intervals of the \textsc{lsig}, that is, its domain and co-domain, we realize the \textsc{nlsig}. This is a more powerful function definition or model of $n$ peak inflection points and $2n$ valley inflection-points, where, for a partition defined by a single peak inflection-point, there are $p$ hyper-parameters to realize almost any kind of smooth growth shape.

Notably, an important characteristic of the \textsc{lsig} when $n=1$ is that at the peak inflection-point, the $y$ is at half of its max-min interval difference on the $y$-dimension. It should be strongly noted that this does not in any way imply a symmetric growth pattern, except in some cases, when the max-min interval of $\mathrm{x}$ is equally partitioned.

\begin{figure}[]
	\centering
	\includegraphics[width=0.8\linewidth]{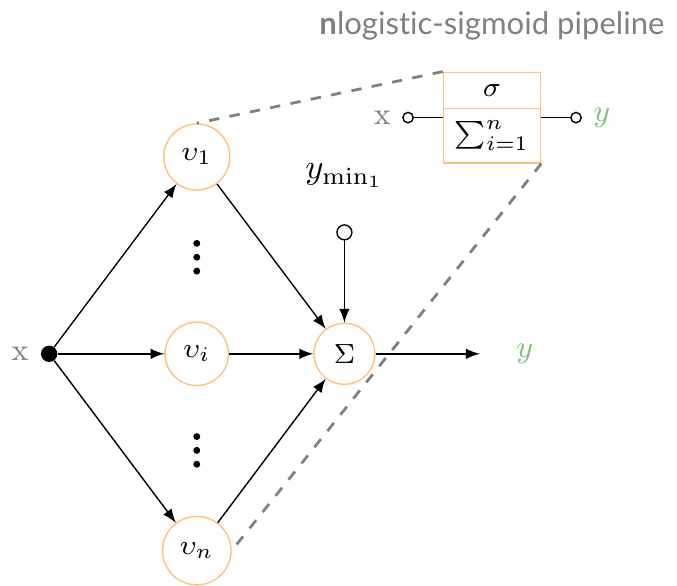}
	\caption[Core \textsc{nlsig} Pipeline]{Core SISO \textsc{nlsig} Pipeline}
	\label{fig:nlsigsiso}
\end{figure}

\begin{figure}[]
	\centering
	\includegraphics[width=0.8\linewidth]{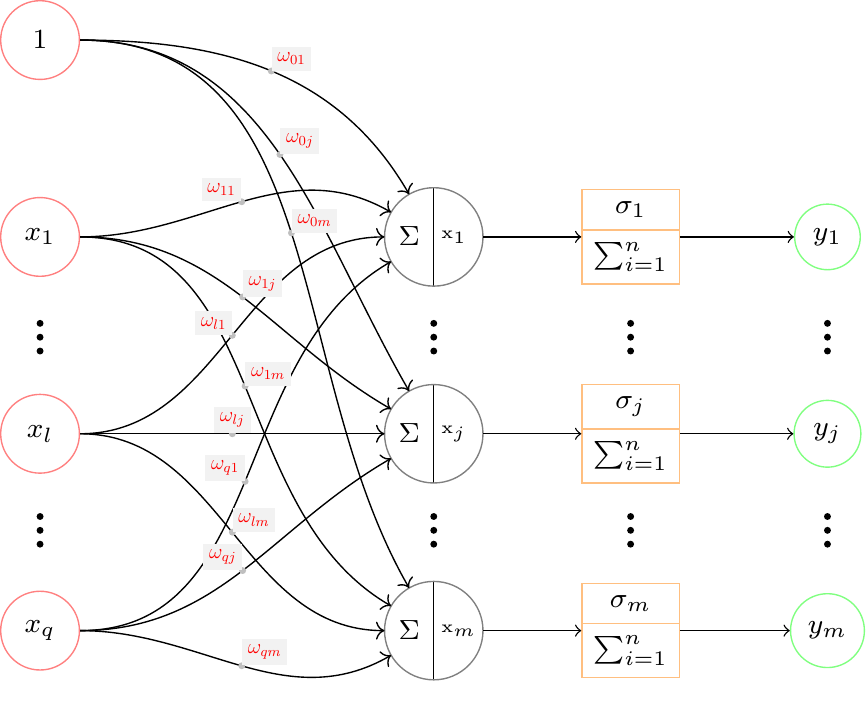}
	\caption[MIMO \textsc{nlsig} Network]{General MIMO \textsc{nlsig} Network}
	\label{fig:nlsigmimo}
\end{figure}

Consequently, we provide formal, compact and unifying definition of the \textsc{nlsig} pipeline for any arbitrary finite interval in single-output and multiple-output forms in \Crefrange{sec_sisodef}{sec_mimodef} respectively. The core pipeline, the single-input single-output or multiple-input single-output (SISO/MISO) form is illustrated in \Cref{fig:nlsigsiso}. The general multiple-input multiple-output (MIMO) network is illustrated in \Cref{fig:nlsigmimo}. We also provide definitions for the derivatives, jacobian and hessian of the function for use with gradient-based optimization methods. Also, the logistic metrics are introduced in \Cref{sec_xiryir}.

\section{SISO/MISO Pipeline}\label{sec_sisodef}
In contrast to the classic logistic definition, a repeating sigmoid behaviour found in the motion-profile of ant-swarms can be emulated within any finite input-output universe specified by $[x_{\min}, x_{\max}]$ and $[y_{\min}, y_{\max}]$. The specified finite input and output universe (space) when partitioned into $n$ equal sub-spaces with $n$ inflection points at $x=\delta_i$ leads to $\upsilon_i$ sub-sigmoids, where $i=1,...,n$. This transforms the classic logistic-sigmoid function to a sum of $n$ shifted sigmoids each with a bounded and continuous gradient map over the surface of the input-output universe on an exponential base $b_i$, where $b_i>1 \in \mathbb{R}$ but can be fixed to a standard integer base such as bianry $2$, decimal $10$, vigesimal $20$, or instead the real-valued natural exponential $e$. The parameter $n$ defines the number of sub-sigmoids in the function, $\upsilon_i$ is the sub-output representing the $i$th sub-sigmoid.

Accordingly, the single-output \textsc{nlsig} denoted $\text{\textsc{nlsig}}\pm\colon\mathbb{R}^q\mapsto\mathbb{R}^1$, where $n\in\mathbb{N}^+$
is defined as follows:
\begin{defn}\label{siodef}	
	 Given an input $\mathrm{x} = w_0 + \sum_{l = 1}^{q} w_lx_l$, such that $x\in \mathbb{R}^q$, $q \geq 1$ and $\mathrm{x} \in\mathbb{R}^1$, the odd output of the \textsc{nlsig} is $y = \sigma (\mathrm{x})\in \mathbb{R}^1$, the even output is $g = \sigma^{\prime} (\mathrm{x})\in \mathbb{R}^1$, with $2 \le p \le 7$ number of hyper-parameters for each of the $i$ partitions:
	 \begin{IEEEeqnarray}{rCl}
	 	y &=& y_{\min_1} + \sum_{i = 1}^{n} \si{\upsilon}_i\\
	 	g &=& \dv{y}{\mathrm{x}} = \sum_{i = 1}^{n} \dv{\upsilon_i}{\mathrm{x}} =  \pm \sum_{i = 1}^{n} \alpha_i\ln(b_i)\upsilon_i\left( \frac{\upsilon_i}{\Delta y_i} - 1 \right)\\
	 	h &=& \dv[2]{y}{\mathrm{x}} = \sum_{i = 1}^{n} \dv[2]{\upsilon_i}{\mathrm{x}}\IEEEnonumber\\
	 	 &= &\pm \sum_{i = 1}^{n} \alpha_i\ln(b_i)\dv{\upsilon_i}{\mathrm{x}}\left( \frac{2\upsilon_i}{\Delta y_i} - 1 \right) \in \mathbb{R}^1	 		 	
	 \end{IEEEeqnarray}
	 where the arbitrary sub-intervals for each $i$-th sub-sigmoid partitions are defined as:
	 \begin{IEEEeqnarray}{rCl}
	 	\Delta \mathrm{x}_i &=& \mathrm{x}_{\max_i} - \mathrm{x}_{\min_i}\\
	 	\Delta y_i &=& y_{\max_i} - y_{\min_i}
	 \end{IEEEeqnarray}
 	and the intermediate parameters are:
	 \begin{IEEEeqnarray}{rCl}
		\upsilon_i  &=& \frac{\Delta y_i}{D_i}\\
		D_i &=& 1 + b_{i}^{\pm \alpha_i {}_i(\mathrm{x} - \delta_i)}\\
		\alpha_i  &=& \frac{2\lambda_i}{\Delta x_i}
	\end{IEEEeqnarray} 	
 	The first and second-order derivatives with respect to the input variables $x_l$ are:
	\begin{IEEEeqnarray}{rCl}
	g_l &=& \dv{y}{x_l} = w_lg\\
	h_l &=& \dv[2]{y}{x_l} = w_lh
	\end{IEEEeqnarray}
\end{defn}		

Further, given a number of real data-points $r_d$, such that, $ 1 \le d \le \mathrm{D}$, often, the objective is to find $y_d$ that minimizes in the least-squares sense, an error objective of the form: $E = \frac{1}{2} \sum_{d = 1}^{D} {e_d^2}$, where $e_d = r_d - y_d$.

\begin{defn}
	At the logistic input-output layer, the first-order partial derivatives of the output $y$ to the hyper-parameters can be given as: let $t_i=\pm \ln(b_{i})\upsilon_i\left( \pdv{y}{y_{\max_i}} - 1 \right)$, then
	\begin{IEEEeqnarray}{cCl}
	\pdv{y}{y_{\max_i}} &=& \frac{\upsilon_i}{\Delta y_i}\\
	\pdv{y}{y_{\min_i}} &=& 1 - \pdv{y}{y_{\max_i}}\\
	\pdv{y}{\delta_i} &=& -\alpha_i t_i\\
	\pdv{y}{\alpha_i} &=& (\mathrm{x} - \delta_i)t_i\\
	\pdv{y}{\lambda_i} &=& \frac{2}{\Delta \mathrm{x}_i}\pdv{y}{\alpha_i}\\
	\pdv{y}{\mathrm{x}_{\max_i}} &=& -\frac{\alpha_i}{\Delta \mathrm{x}_i}\pdv{y}{\alpha_i}\\
	\pdv{y}{\mathrm{x}_{\min_i}} &=& \frac{\alpha_i}{\Delta \mathrm{x}_i}\pdv{y}{\alpha_i}\\
	\pdv{y}{b_i} &=& \frac{\alpha_i}{b_i\ln(b_i)}\pdv{y}{\alpha_i}
	\end{IEEEeqnarray}	
	
	The jacobian and hessian matrix, where $1 \le i \le n$  are:
	\begin{IEEEeqnarray}{c}
		\mathbf{j}^{i} = \sum_{d=1}^{D} {e_d\,\mathbf{j}_d^{i,y}}\\
		\mathbf{H}^{i} = \left(\mathbf{j}^{i}\right)^\mathsf{T}\,\mathbf{j}^{i}  \in \mathbb{R}^{p \times p}
	\end{IEEEeqnarray}
	where,
	\begin{IEEEeqnarray}{c}
	\mathbf{j}^{i,y} = \sum_{d=1}^{D} {\mathbf{j}_{d}^{i,y}}\\
	\mathbf{j}_{d}^{i,y} = \left[\begin{array}{cl}
		\pdv{y}{b_i}, \pdv{y}{\lambda_i},  \pdv{y}{\mathrm{x}_{\max_i}}, \pdv{y}{\mathrm{x}_{\min_i}},\pdv{y}{\delta_i},
	\end{array}\right.\IEEEnonumber\\
	\left. \begin{array}{lr}
		\pdv{y}{y_{\max_i}},\pdv{y}{y_{\min_i}}
	\end{array} \right] \in \mathbb{R}^{1 \times p}
	\end{IEEEeqnarray}
	
	\end{defn}

\begin{defn}	
	At the weighted-input layer, the definitions for the partial-derivatives are:
	\begin{IEEEeqnarray}{c}
		\pdv{E}{y} = e_j\\
		\pdv{E}{\mathrm{x}} = e\,\dv{y}{\mathrm{x}}\\
		\pdv{y}{\omega_l} =\left\{ \begin{array}{cl}
			\dv{y}{\mathrm{x}} & \mbox{if } l = 0\\
			& \\
			x_l\,\dv{y}{\mathrm{x}} & \mbox{if } l > 0
		\end{array} \right.\\
		\pdv{E}{\omega_l} = e\,\pdv{y}{\omega_l}
	\end{IEEEeqnarray}
	such that the jacobian matrix, hessian matrix, and gradient vector are respectively:
	\begin{IEEEeqnarray}{c}
	\mathbf{j}^w = \sum_{d=1}^{D} {\mathbf{j}_d^w}\\
	\mathbf{H}^w = {\left(\mathbf{j}^{w}\right)}^\mathsf{T}\, \mathbf{j}^w\\
	\mathbf{g}^w  = \sum_{d=1}^{D} {{\left(\mathbf{j}_d^{w}\right)}^\mathsf{T}\, e_d}
	\end{IEEEeqnarray}
	where,
	\begin{IEEEeqnarray}{c}
		\mathbf{j}_d^w = \left[\pdv{E}{\omega_0},\dots,\pdv{E}{\omega_l},\dots\pdv{E}{\omega_q} \right] \in \mathbb{R}^{1 \times {(q+1)}}
	\end{IEEEeqnarray}		
	
\end{defn}
\begin{rem}
	The single-output \textsc{nlsig} model can be used for binary logistic regression, where the $y_{\max_n}=1$ and $y_{\min_n}=0$. The weighted sum of inputs is $\mathrm{x}$ and called the logit or quantile. The inputs can be single, that is, univariate (single explanatory input-variables) or multiple, that is: multivariate (multiple explanatory input-variables).
	Regression, loosely speaking is `data fitting'. It involves the use of a set of independent or explanatory $x_l$ variables to predict a set of dependent variables or outcomes $y_j$ through some optimization (learning) method like the non-linear least-squares.
\end{rem}

\section{SIMO/MIMO Pipeline}\label{sec_mimodef}

The multiple-output \textsc{nlsig} denoted $\text{\textsc{nlsig}}\pm\colon\mathbb{R}^q\mapsto\mathbb{R}^m$, where $n\in\mathbb{N}^+$, given $q \geq 1$ inputs and $m > 1$ outputs
is defined as follows:
\begin{defn}\label{siodef}	
	Given an input $\mathrm{x}_j = w_{0j} + \sum_{l = 1}^{q} w_{lj}x_l$, such that $x\in \mathbb{R}^q$, $q \geq 1$ and $\mathrm{x} \in\mathbb{R}^m$, $1 \le j \le m$, the odd output of the \textsc{nlsig} is $y = \sigma (\mathrm{x})\in \mathbb{R}^m$, the even output is $g = \sigma^{\prime} (\mathrm{x})\in \mathbb{R}^m$, with $2 \le p \le 7$ number of hyper-parameters for each of the $i$ partitions in the $j$-th pipeline:
	\begin{IEEEeqnarray}{rCl}
	y_j &=& y_{\min_{1j}} + \sum_{i = 1}^{n_j} \upsilon_{ij}\\
	g_j &=& \sigma_{j}^{\prime}(\mathrm{x}) = \dv{y_{j}}{\mathrm{x}_{j}} = \sum_{i = 1}^{n_j} \dv{\upsilon_{ij}}{\mathrm{x}_j} \IEEEnonumber\\
	&=& \pm \sum_{i = 1}^{n_j} \alpha_{ij}\ln(b_{ij})\upsilon_{ij}\left( \frac{\upsilon_{ij}}{\Delta y_{ij}} - 1 \right) \in \mathbb{R}^{1}\\
	h_j &=& \sigma_{j}^{\prime\prime}(\mathrm{x}) = \dv[2]{y_j}{\mathrm{x}_j} = \sum_{i = 1}^{n_j} \dv[2]{\upsilon_{ij}}{\mathrm{x}_j} \IEEEnonumber\\
	&=& \pm \sum_{i = 1}^{n_j} \alpha_{ij}\ln(b_{ij})\dv{\upsilon_{ij}}{\mathrm{x}_j}\left( \frac{2\upsilon_{ij}}{\Delta y_{ij}} - 1 \right) \in \mathbb{R}^{1}	 		 	
	\end{IEEEeqnarray}
	where the arbitrary sub-intervals for each $i$-th sub-sigmoid partitions for the $j$-th output are defined as:
	\begin{IEEEeqnarray}{rCl}
	\Delta \mathrm{x}_{ij} = \mathrm{x}_{\max_{ij}} - \mathrm{x}_{\min_{ij}}\\
	\Delta y_{ij} = y_{\max_{ij}} - y_{\min_{ij}}
	\end{IEEEeqnarray}
	and the intermediate parameters are:
	\begin{IEEEeqnarray}{rCl}
	\upsilon_{ij} = \frac{\Delta y_{ij}}{D_{ij}}\\
	D_{ij} = 1 + b_{ij}^{\pm \alpha_{ij}(\mathrm{x}_j - \delta_{ij})}\\
		\alpha_{ij} = \frac{2\lambda_{ij}}{\Delta \mathrm{x}_{ij}}
	\end{IEEEeqnarray} 	
	for the multinomial classification case, like the softmax, where the expectation is that the $\sum_{j=1}^{m} y_j = y_{\max_{nj}}$, then $D_{ij}$ becomes
	\begin{IEEEeqnarray}{c}
	D_{ij} = 1 + \left(\sum_{k = 1,  k \neq j}^{m} b_{ij}^{\pm  (\alpha_{ij}(\mathrm{x}_j - \delta_{ij}) - \alpha_{ik}(\mathrm{x}_k - \delta_{ik}))} \right)
	\end{IEEEeqnarray}
	The first and second-order derivatives with respect to the input variables $x_l$ are:
	\begin{IEEEeqnarray}{rCl}
		g_{lj} &=& \dv{y_j}{x_{lj}} = w_{lj}\,g_l\\
		h_{lj} &=& \dv[2]{y_j}{x_{lj}} = w_{lj}\,h_l
	\end{IEEEeqnarray}
\end{defn}		

Further, given a number of real data-points $r_d$, such that, $ 1 \le d \le \mathrm{D}$, often, the objective is to find $y_d$ that minimizes in the least-squares sense, an error objective of the form: $E = \frac{1}{2} \sum_{d = 1}^{D}  \sum_{j = 1}^{m} {e_{jd}^2}$, where $e_{jd} = r_{jd} - y_{jd}$.

\begin{defn}
	At the logistic input--output layer, the first-order partial derivatives of the output $y$ to the hyper-parameters can be given as: let $t_{ij} =  \pm \ln(b_{ij})\upsilon_{ij}\left( \pdv{y_j}{y_{\max_{ij}}} - 1 \right)$, then
	\begin{IEEEeqnarray}{cCl}
		\pdv{y_j}{y_{\max_{ij}}} &=& \frac{\upsilon_{ij}}{\Delta y_{ij}}\\
		\pdv{y_j}{y_{\min_{ij}}} &=& 1 - \pdv{y_j}{y_{\max_{ij}}}\\
		\pdv{y_j}{\delta_{ij}} &=& -\alpha_{ij} t_{ij}\\
		\pdv{y_j}{\alpha_{ij}} &=& (\mathrm{x}_j - \delta_{ij})t_{ij}\\
		\pdv{y_j}{\lambda_{ij}} &=& \frac{2}{\Delta \mathrm{x}_{ij}}\pdv{y_j}{\alpha_{ij}}\\
		\pdv{y_j}{\mathrm{x}_{\max_{ij}}} &=& -\frac{\alpha_{ij}}{\Delta \mathrm{x}_{ij}}\pdv{y_j}{\alpha_{ij}}\\
		\pdv{y_j}{\mathrm{x}_{\min_{ij}}} &=& \frac{\alpha_{ij}}{\Delta \mathrm{x}_{ij}}\pdv{y_j}{\alpha_{ij}}\\
		\pdv{y_j}{b_{ij}} &=& \frac{\alpha_{ij}}{b_{ij}\ln(b_{ij})}\pdv{y_j}{\alpha_{ij}}
	\end{IEEEeqnarray}	
	
	The jacobian and hessian matrix, where $1 \le i \le n$ for the $j$-th output are:
	\begin{IEEEeqnarray}{c}
		\mathbf{j}_j^{i,y} = \sum_{d=1}^{D} {\mathbf{j}_{jd}^{i,y}}\\
		\mathbf{H}_j^{i}  = \left(\mathbf{j}_j^{i}\right)^\mathsf{T}\,\mathbf{j}_j^{i}  \in \mathbb{R}^{p \times p}
	\end{IEEEeqnarray}
	where,
	\begin{IEEEeqnarray}{c}
		\mathbf{j}_j^{i,y} = \sum_{d=1}^{D} {\mathbf{j}_{jd}^{i,y}}\\
		\mathbf{j}_{d}^{i,y} = \left[\begin{array}{cl}
			\pdv{y_j}{b_{ij}}, \pdv{y_j}{\lambda_{ij}}, \pdv{y_j}{\mathrm{x}_{\max_{ij}}}, \pdv{y_j}{\mathrm{x}_{\min_{ij}}}, \pdv{y_j}{\delta_{ij}},
		\end{array}\right.\IEEEnonumber\\
		\left. \begin{array}{lr}
			\pdv{y_j}{y_{\max_{ij}}}, \pdv{y_j}{y_{\min_{ij}}}
		\end{array} \right] \in \mathbb{R}^{1 \times p}
	\end{IEEEeqnarray}
	
\end{defn}

\begin{defn}	
	At the weighted-input layer, the definitions for the partial-derivatives are:
	\begin{IEEEeqnarray}{c}
		\pdv{E}{y_{j}} = e_j\\
		\pdv{E}{\mathrm{x}_{j}} = e_j\,\dv{y_j}{\mathrm{x}_{j}}\\
		\pdv{y_j}{\omega_{lj}} =\left\{ \begin{array}{cl}
			\dv{y_j}{\mathrm{x}_j} & \mbox{if } l = 0\\
			& \\
			x_l\,\dv{y_j}{\mathrm{x}_j} & \mbox{if } l > 0
		\end{array} \right.\\
	\pdv{E}{\omega_{lj}} = e_j\,\pdv{y_j}{\omega_{lj}}
	\end{IEEEeqnarray}
	such that the jacobian matrix, hessian matrix, and gradient vector are respectively:
	\begin{IEEEeqnarray}{c}
		\mathbf{j}_j^w = \sum_{d=1}^{D} {\mathbf{j}_{jd}^w}\\
		\mathbf{H}_j^w = {\left(\mathbf{j}_j^{w}\right)}^\mathsf{T}\, \mathbf{j}_j^w\\
		\mathbf{g}_j^w  = \sum_{d=1}^{D} {{\left(\mathbf{j}_{jd}^{w}\right)}^\mathsf{T}\, e_{jd}}
	\end{IEEEeqnarray}
	where,
	\begin{IEEEeqnarray}{c}
	\mathbf{j}_{jd}^w = \left[\pdv{E}{\omega_{0j}},\dots,\pdv{E}{\omega_{lj}},\dots\pdv{E}{\omega_{qj}} \right] \in \mathbb{R}^{1 \times {(q+1)}}
	\end{IEEEeqnarray}		
	
\end{defn}

\section{Logistic-growth Metrics}\label{sec_xiryir}

\begin{defn}[Y-variable to Inflection Ratio]
	The $\mathrm{YIR}$ is useful for indicating the state of the rate of incident increase or decrease over the curve's interval at a particular cumulative value. The theoretical idea is that at the $y$ corresponding to peak inflection points, the $\mathrm{YIR}$ value is always $0.5$ for the logistic curve.
\begin{IEEEeqnarray}{c}
\mathrm{YIR} = \frac{y - y_{\min_i}}{y_{\max_i} - y_{\min_i}}
\end{IEEEeqnarray}
such that for a growing or decaying process:
\begin{IEEEeqnarray}{c}
	\begin{cases}
		\mathrm{YIR} < 0.5 & \mbox{if } \text{generally increasing} \\
		\mathrm{YIR} \approx 0.5 & \mbox{if } \text{generally around a peak value} \\
		\mathrm{YIR} > 0.5 & \mbox{if } \text{generally reducing}
	\end{cases}
\end{IEEEeqnarray}

\end{defn}

\begin{defn}[X-variable to Inflection Ratio]
	The $\mathrm{XIR}$ is useful for indicating how far a certain $\mathrm{x}$ (e.g: time) is from the peak-inflection $\mathrm{x}=\delta_i$.
	\begin{IEEEeqnarray}{c}
		\mathrm{XIR} = b_{i}^{\alpha_i (\mathrm{x} - \delta_i)}
	\end{IEEEeqnarray}
 such that for a growing or decaying process:
	\begin{IEEEeqnarray}{c}
		\begin{cases}
		\mathrm{XIR} < 1 & \mbox{if } \mbox{pre-peak period}\\
		\mathrm{XIR} \approx 1 & \mbox{if } \mbox{peak period} \\
		\mathrm{XIR} > 1, \mathrm{XIR} \approx 0 & \mbox{if } \mbox{post-peak period}
		\end{cases}
	\end{IEEEeqnarray}
\end{defn}

\begin{rem}
Because of the $n$ in its name, the \textsc{nlsig} is essentially itself a neural function.
The basic SISO/MISO pipeline has $1$ hidden unweighted logistic parameter layer. For the SIMO/MIMO pipeline, that becomes $m$ hidden unweighted logistic parameter layers. When explanatory inputs are used, there is an added $1$ hidden weight layer for the SISO/MISO case, and, $m$ hidden weight layers for the SIMO/MIMO case.
Therefore the \textsc{nlsig} is a $2m$ hidden layer neural function. where $m=1$ by default. At the most basic level, for the SISO case, we can assume the weight layer is fixed to unity and the bias weight $\omega_0$ is 0. such that $\mathrm{x} = x_1$. For the remainder of this paper, we shall apply this SISO case to model the COVID-19 growth-process in the world.
\end{rem}

\section{Modelling COVID-19}

Modelling COVID-19 data from official sources, assuming it is reliable, is a way to truly monitor the scale and impact of the COVID-19 pandemic phenomena.  In this context, we apply the \textsc{nlsig} to model the COVID-19 pandemic. Further, to estimate the uncertainty of our model estimates, we
use a parametric bootstrap approach to estimate the 95\% confidence intervals on the uncertainty in the best-fit parametric solution of the \textsc{nlsig} model for data-sets obtained from the World Health Organization up to the 6th of December 2020.

Therefore, given such real-world growth-process described by its noisy data-sets, we developed a problem-based optimization workflow in MATLAB 2019b, as follows, for fitting the COVID-19 datasets to the \textsc{nlsig} function.:
\begin{enumerate}
	\item find the possible locations of the inflection-points in the data, noting that the logistic-sigmoid curve at its core is characterized by three phases: pre-peak, peak and post-peak phases.
	\item use the detected $3n$ inflection-points to find $n$ and use as initial guess to a suitable optimization solver to find an estimated solution to the inflection points and other three hyper-parameters.
	\item applying the bootstrap method, find the 95\% confidence intervals on the best-fit solution of the $p$ hyper-parameters for each $i$-partitions of the logistic layer, obtained as a whole, by the solver.
\end{enumerate}
We illustrate our results with the projected logistic-metrics on the growth-evolution of the COVID-19 pandemic for the world as a whole, and then selected countries in four continental locations of the world.
For Europe, the locales are: UK, Italy, France, Sweden, Turkey, Russia. For Asia, the locales are: China, Japan, South-Korea, India, Israel, Iran, UAE. For America and Australia, the locales are: USA, Canada, Australia, Cuba, Mexico, Brazil. For Africa, the locales are: Nigeria, Ghana, Egypt, South-Africa, Zimbabwe, Kenya.

\begin{figure}[t]
	\centering
	\subfloat[]{\includegraphics[width=0.5\linewidth]{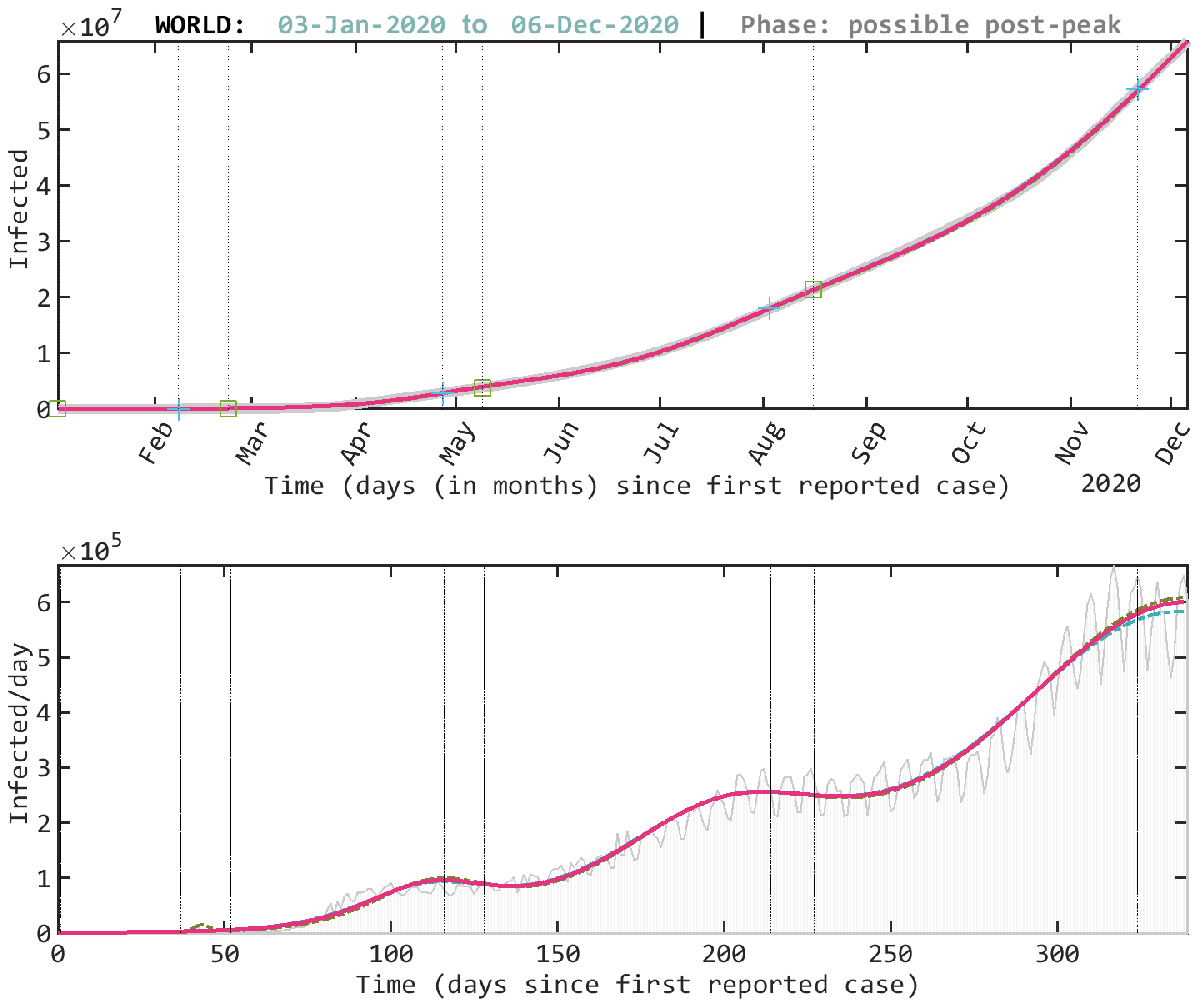}%
	\label{fig:wdi}}\quad
	\subfloat[]{\includegraphics[width=0.5\linewidth]{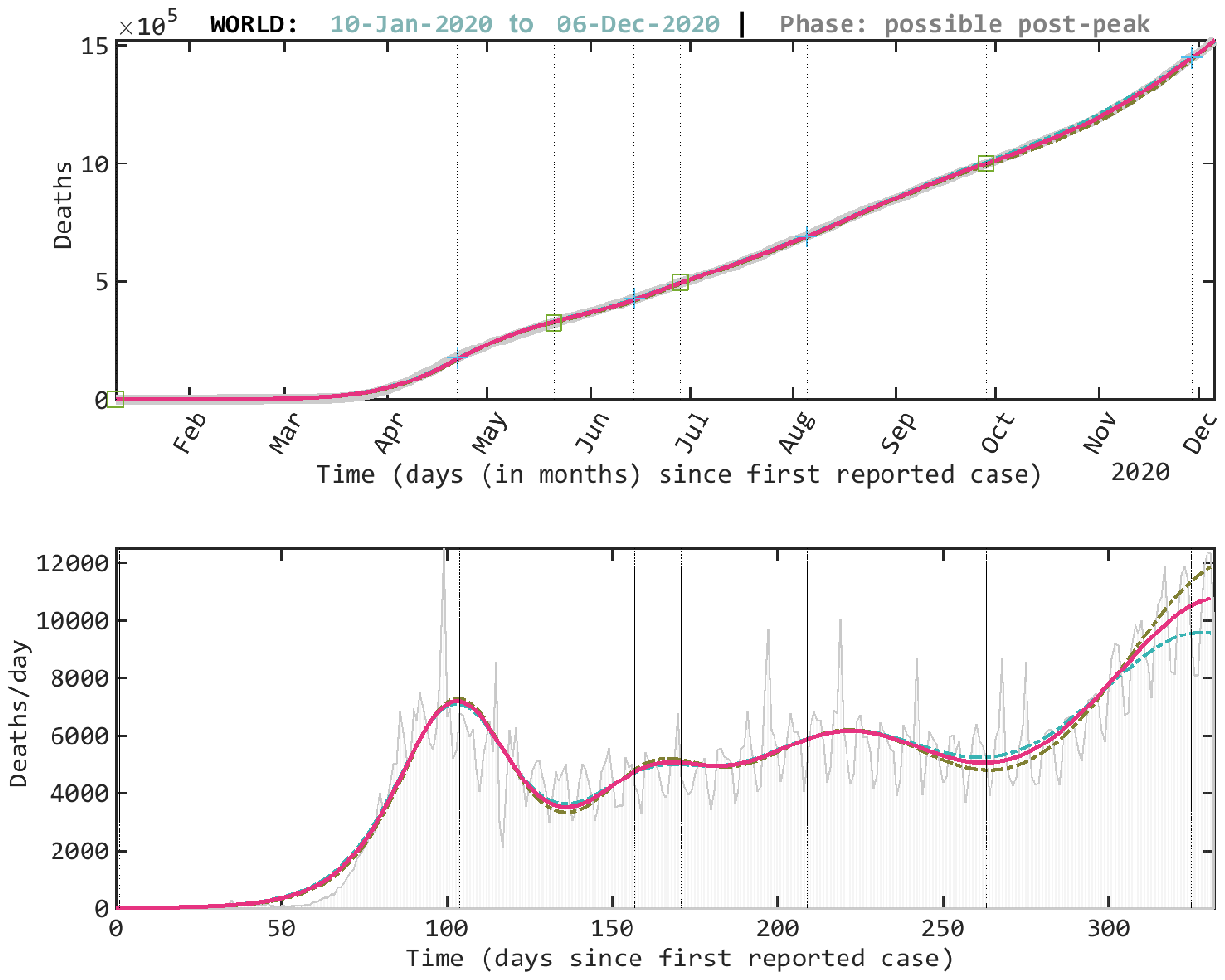}%
	\label{fig:wdd}}
	
	\caption{World (\ref{fig:wdi}) Infections and (\ref{fig:wdd}) Deaths: Growth process described by the nlogistic-sigmoid model (red line), actual COVID-19 data (gray lines and areas). Bootstrapped 95\% upper and lower confidence intervals (dashed brown and dashed blue).}
	\label{fig:wd}
\end{figure}
\begin{figure}[t]
	\centering
	\subfloat[]{\includegraphics[width=0.5\linewidth]{ 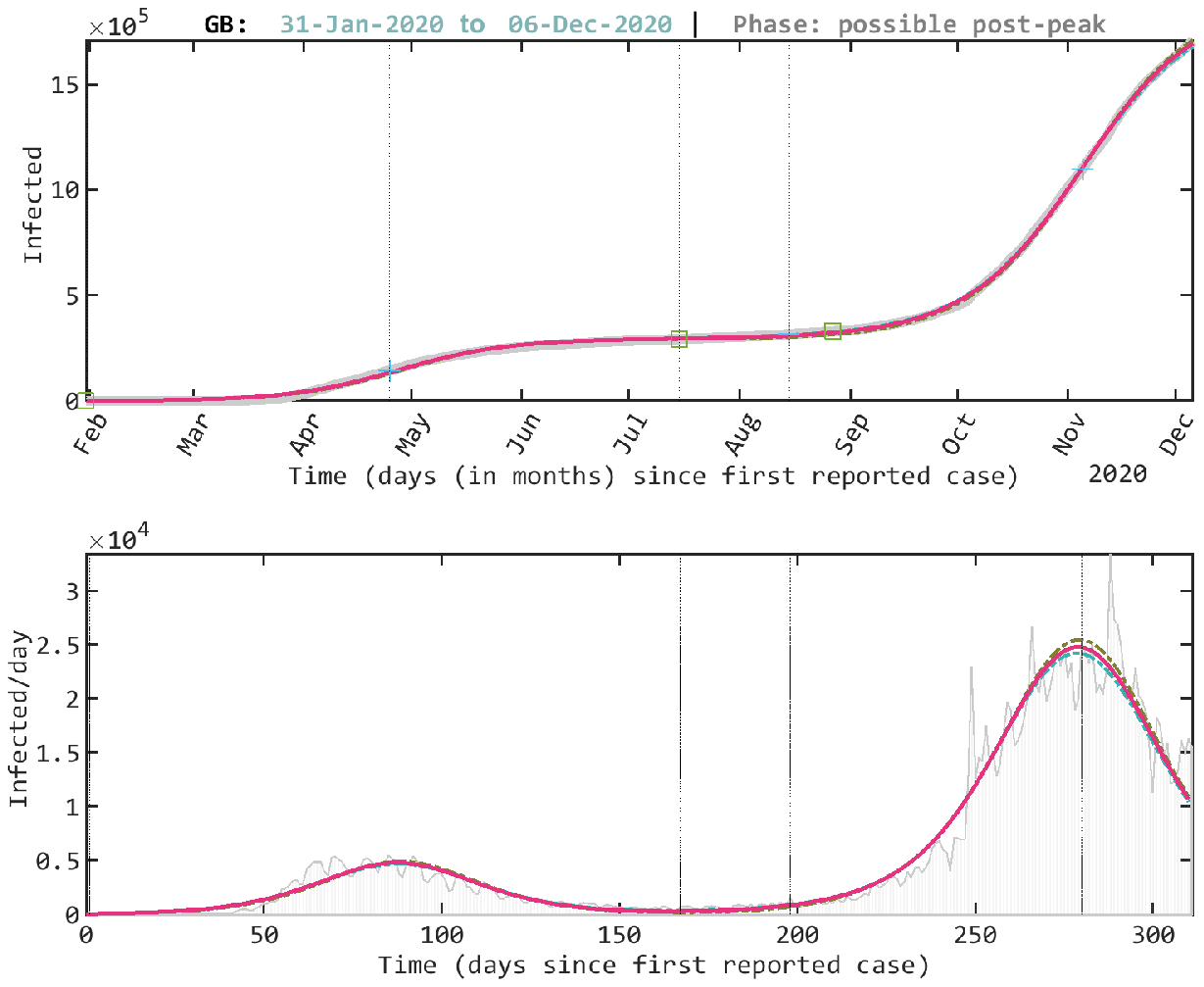}%
		\label{fig:gbi}}\quad
	\subfloat[]{\includegraphics[width=0.5\linewidth]{ 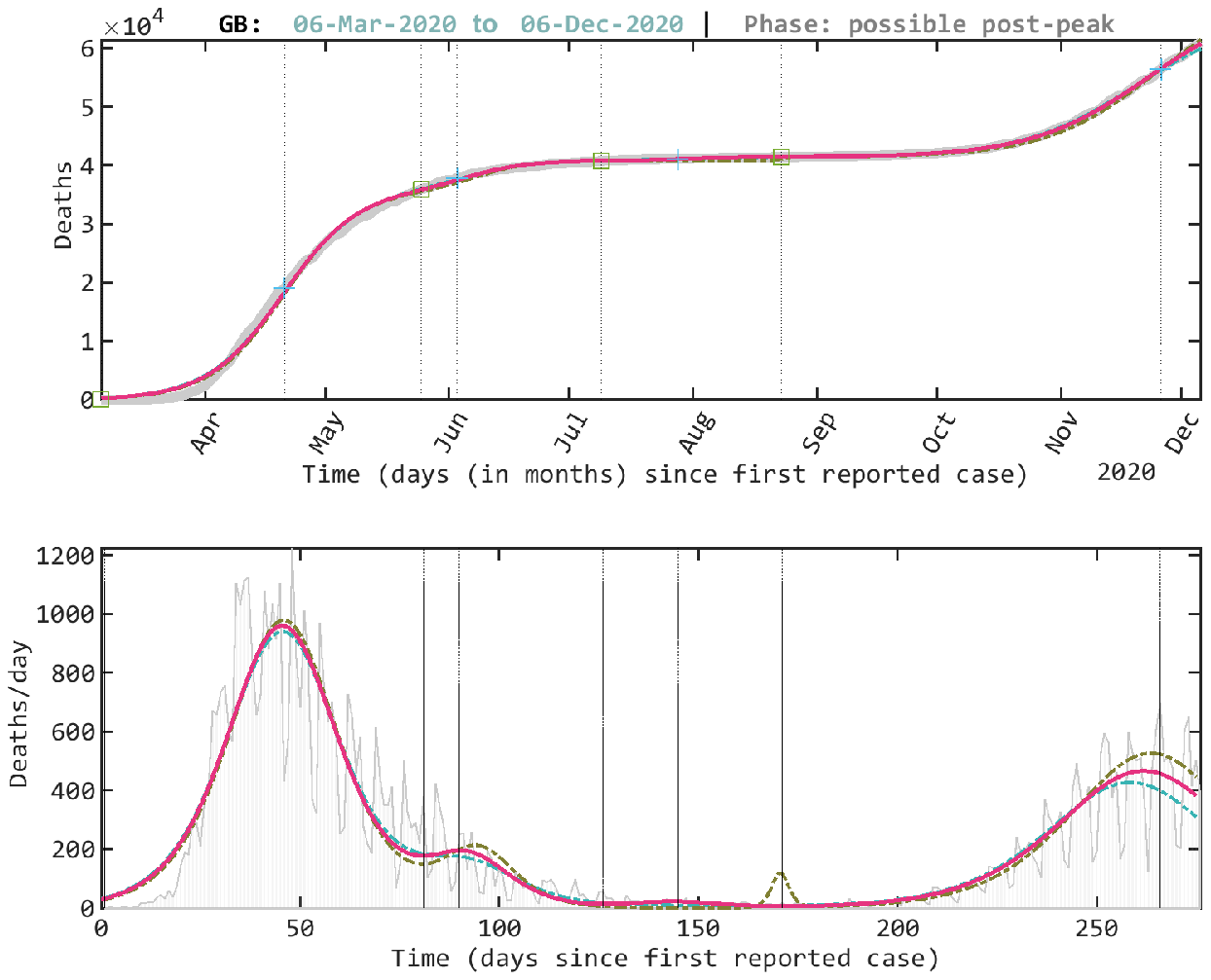}%
		\label{fig:gbd}}
	
	\caption{United Kingdom (\ref{fig:gbi}) Infections and (\ref{fig:gbd}) Deaths: Growth process described by the nlogistic-sigmoid model (red line), actual COVID-19 data (gray lines and areas). Bootstrapped 95\% upper and lower confidence intervals (dashed brown and dashed blue).}
\end{figure}
\begin{figure}[t]
	\centering
	\subfloat[]{\includegraphics[width=0.5\linewidth]{ 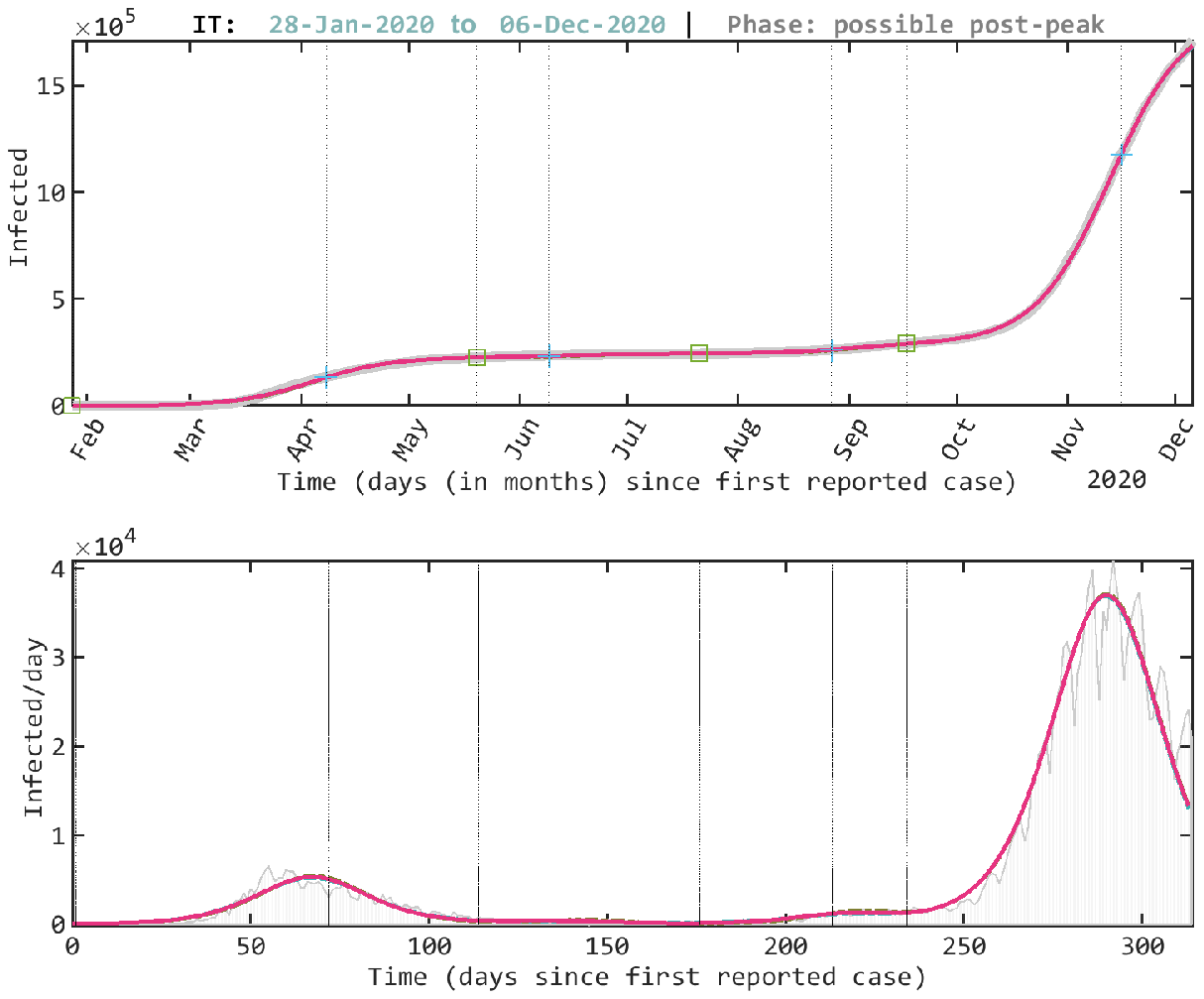}%
		\label{fig:iti}}\quad
	\subfloat[]{\includegraphics[width=0.5\linewidth]{ 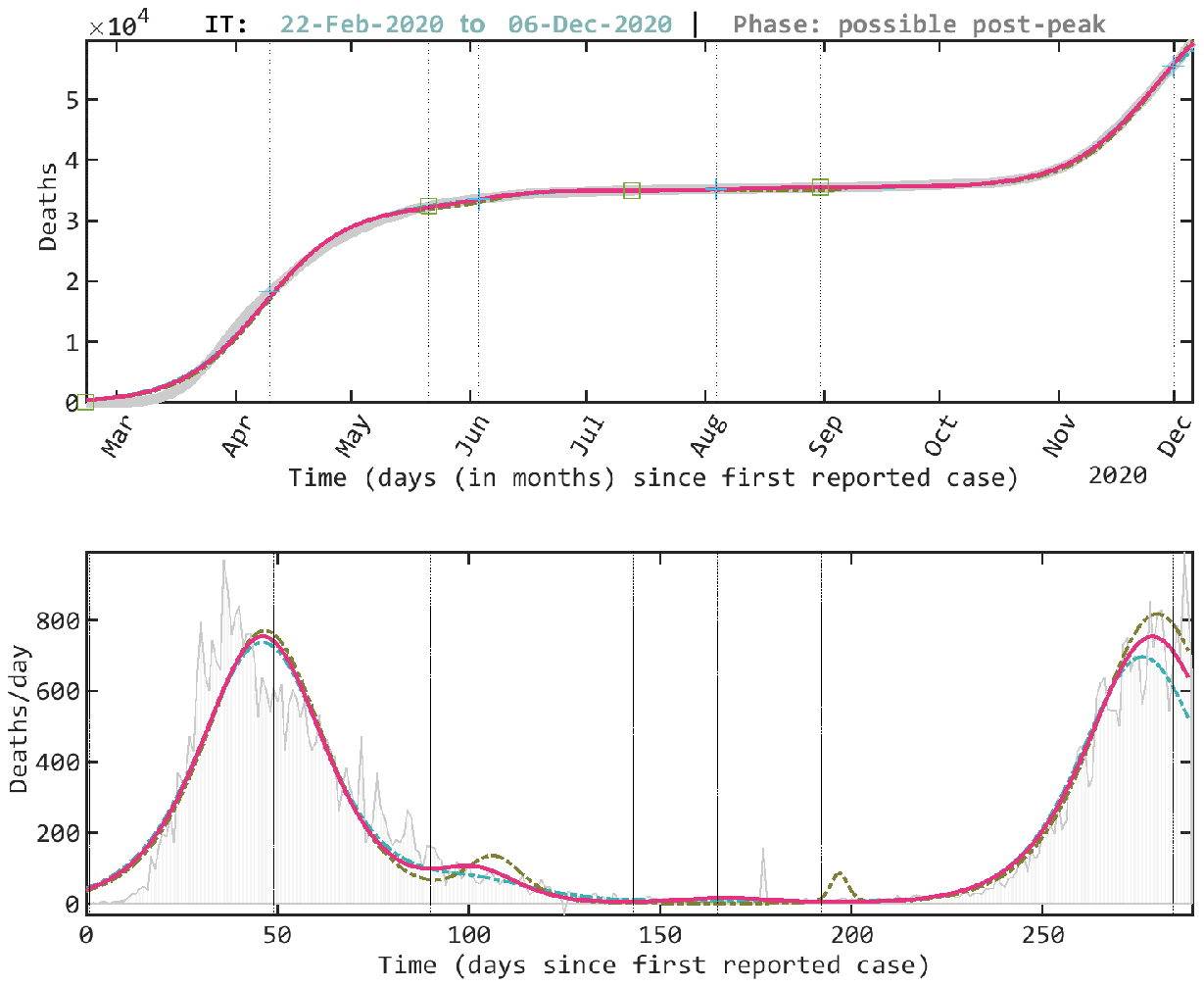}%
		\label{fig:itd}}
	
	\caption{Italy (\ref{fig:iti}) Infections and (\ref{fig:itd}) Deaths: Growth process described by the nlogistic-sigmoid model (red line), actual COVID-19 data (gray lines and areas). Bootstrapped 95\% upper and lower confidence intervals (dashed brown and dashed blue).}
\end{figure}
\begin{figure}[t]
	\centering
	\subfloat[]{\includegraphics[width=0.5\linewidth]{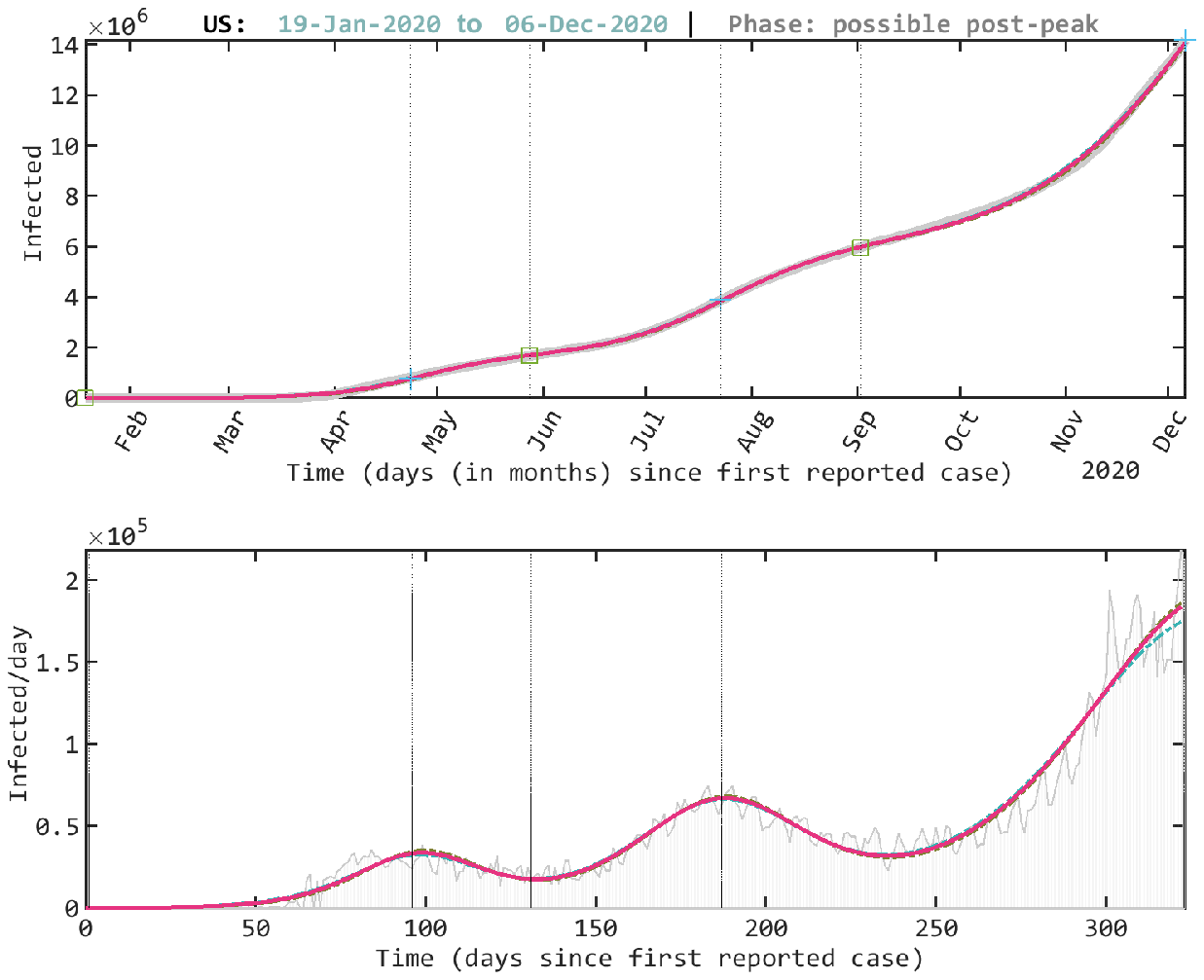}%
		\label{fig:usi}}\quad
	\subfloat[]{\includegraphics[width=0.5\linewidth]{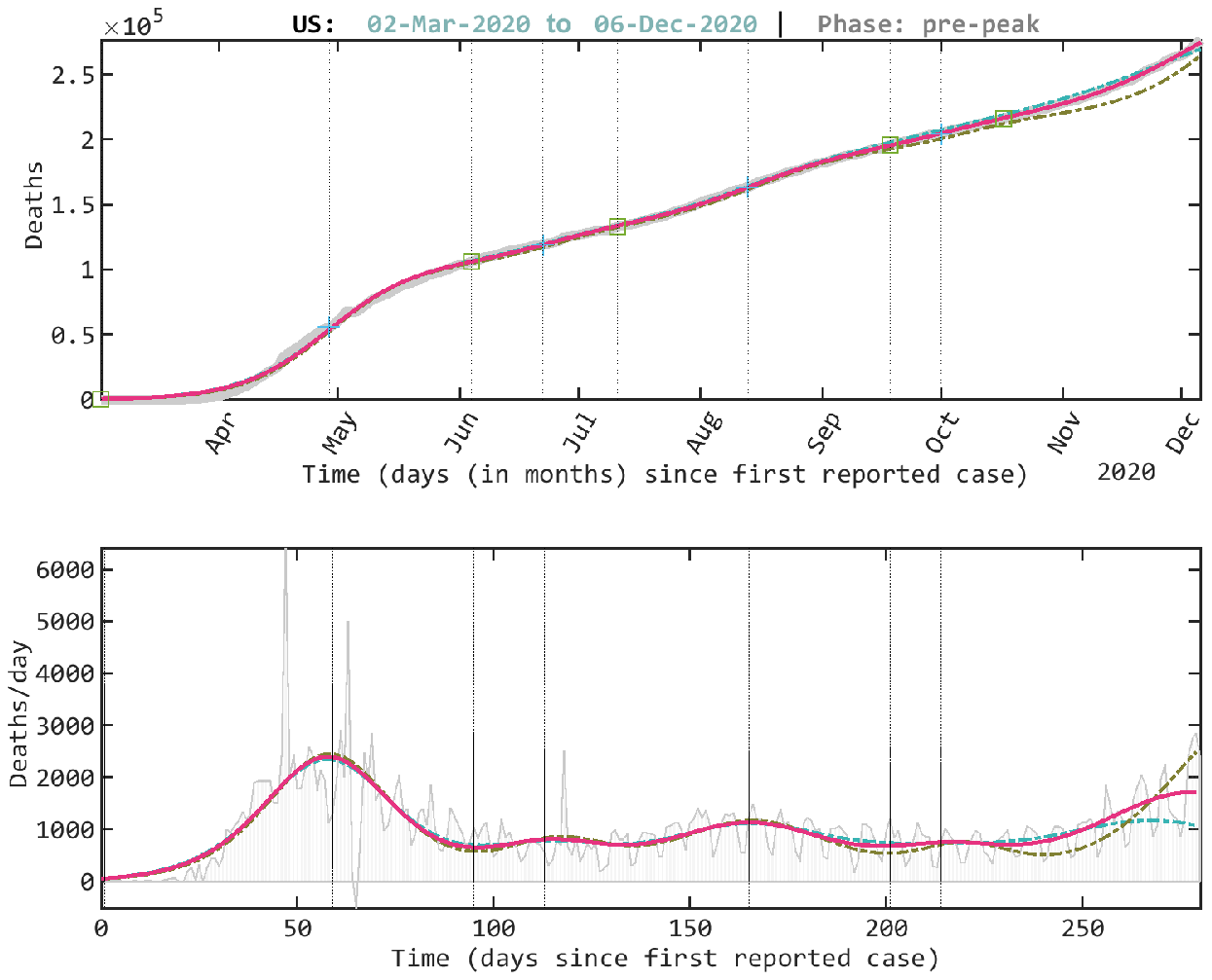}%
		\label{fig:usd}}
	
	\caption{United States of America (\ref{fig:usi}) Infections and (\ref{fig:usd}) Deaths: Growth process described by the nlogistic-sigmoid model (red line), actual COVID-19 data (gray lines and areas). Bootstrapped 95\% upper and lower confidence intervals (dashed brown and dashed blue).}
\end{figure}
\begin{figure}[t]
	\centering
	\subfloat[]{\includegraphics[width=0.5\linewidth]{ 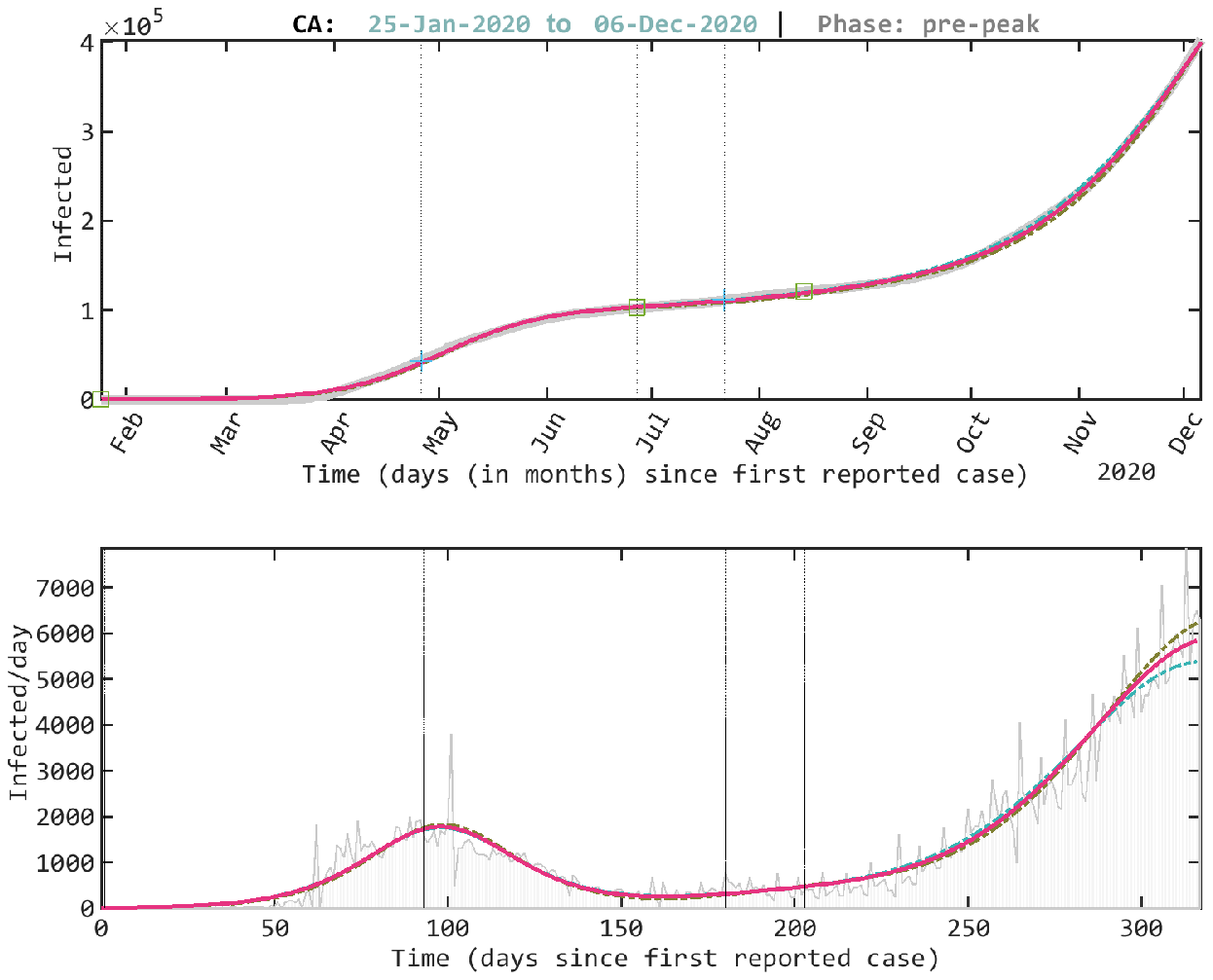}%
		\label{fig:cai}}\quad
	\subfloat[]{\includegraphics[width=0.5\linewidth]{ 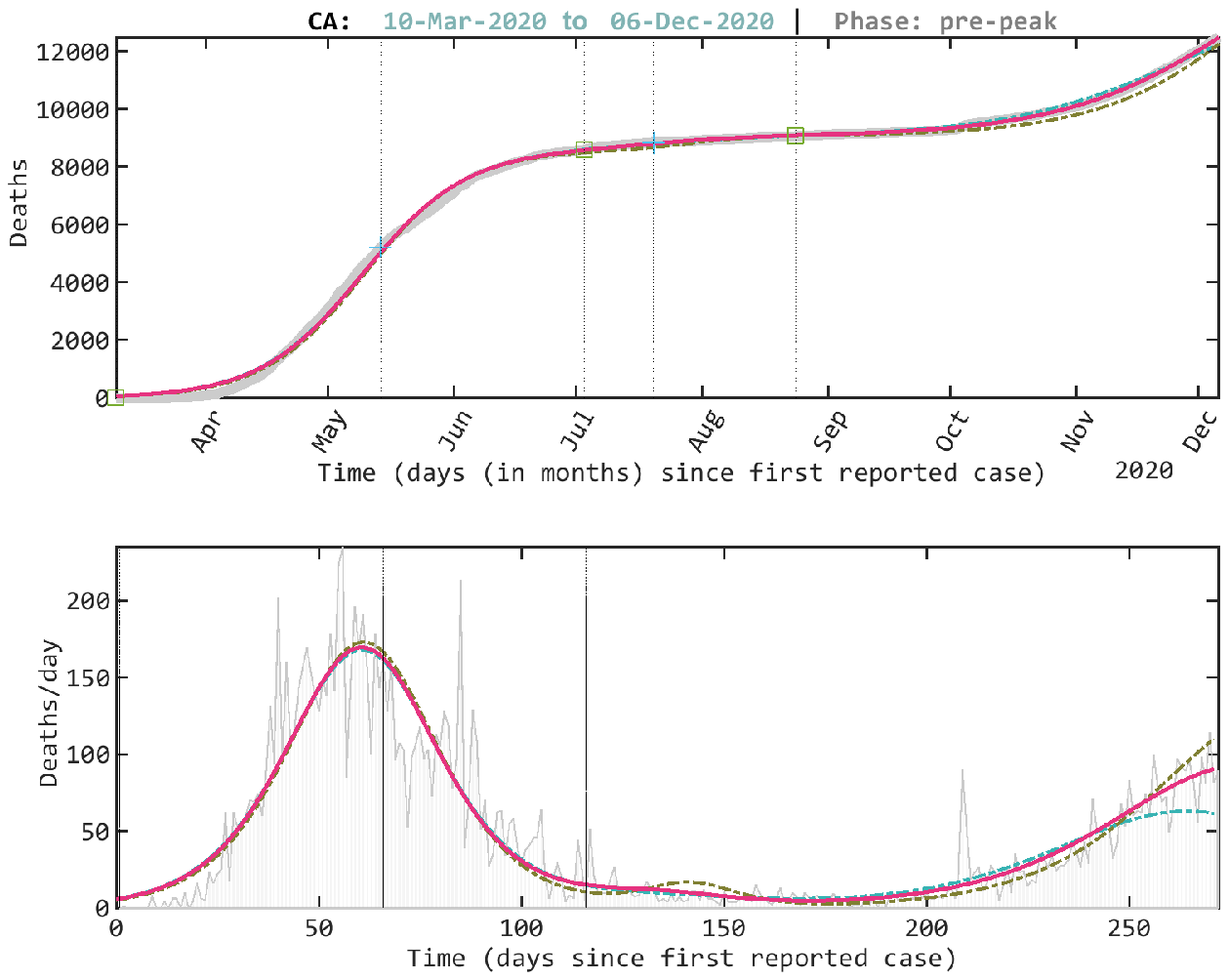}%
		\label{fig:cad}}
	
	\caption{Canada (\ref{fig:cai}) Infections and (\ref{fig:cad}) Deaths: Growth process described by the nlogistic-sigmoid model (red line), actual COVID-19 data (gray lines and areas). Bootstrapped 95\% upper and lower confidence intervals (dashed brown and dashed blue).}
\end{figure}
\begin{figure}[t]
	\centering
	\subfloat[]{\includegraphics[width=0.5\linewidth]{ 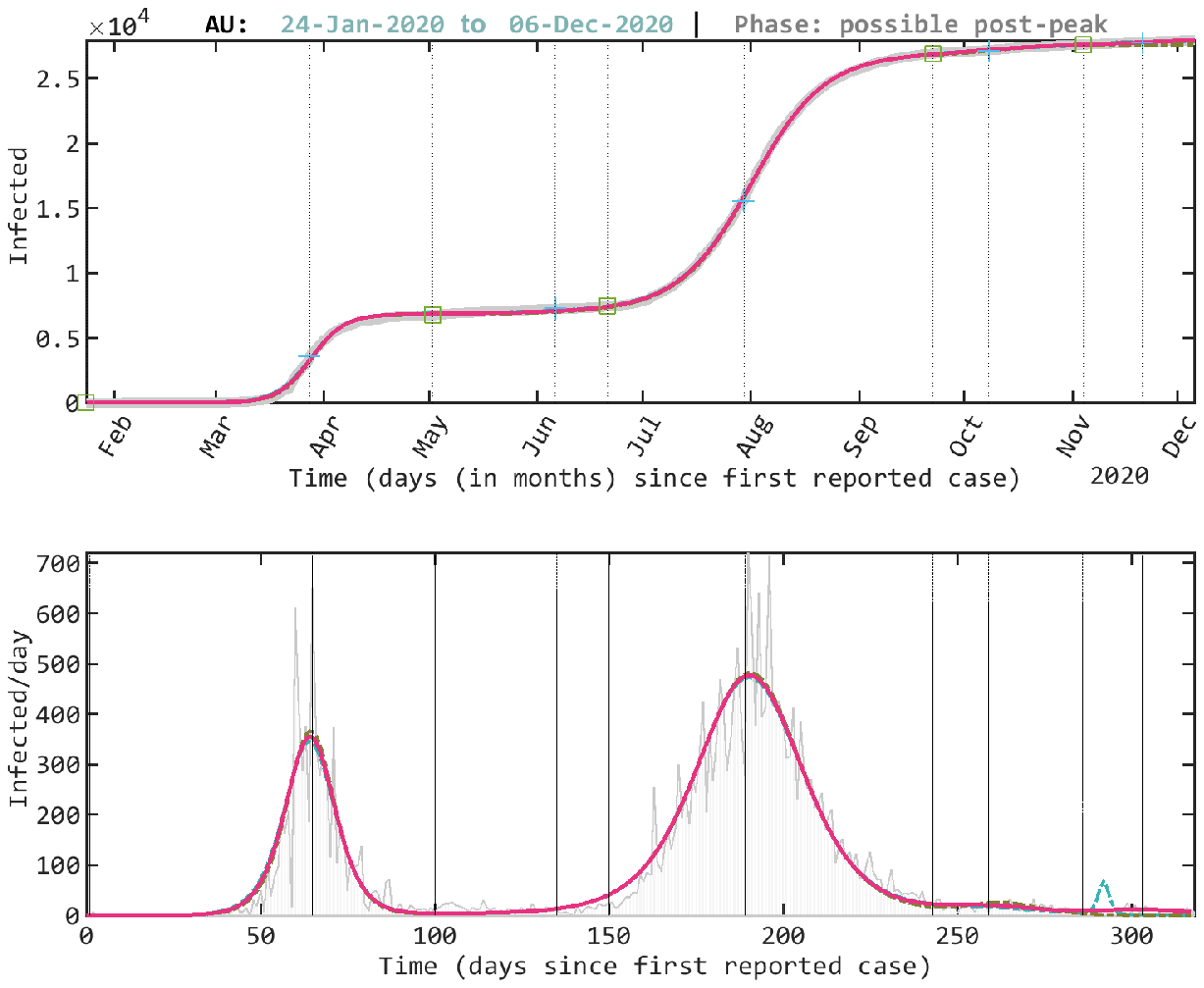}%
		\label{fig:aui}}\quad
	\subfloat[]{\includegraphics[width=0.5\linewidth]{ 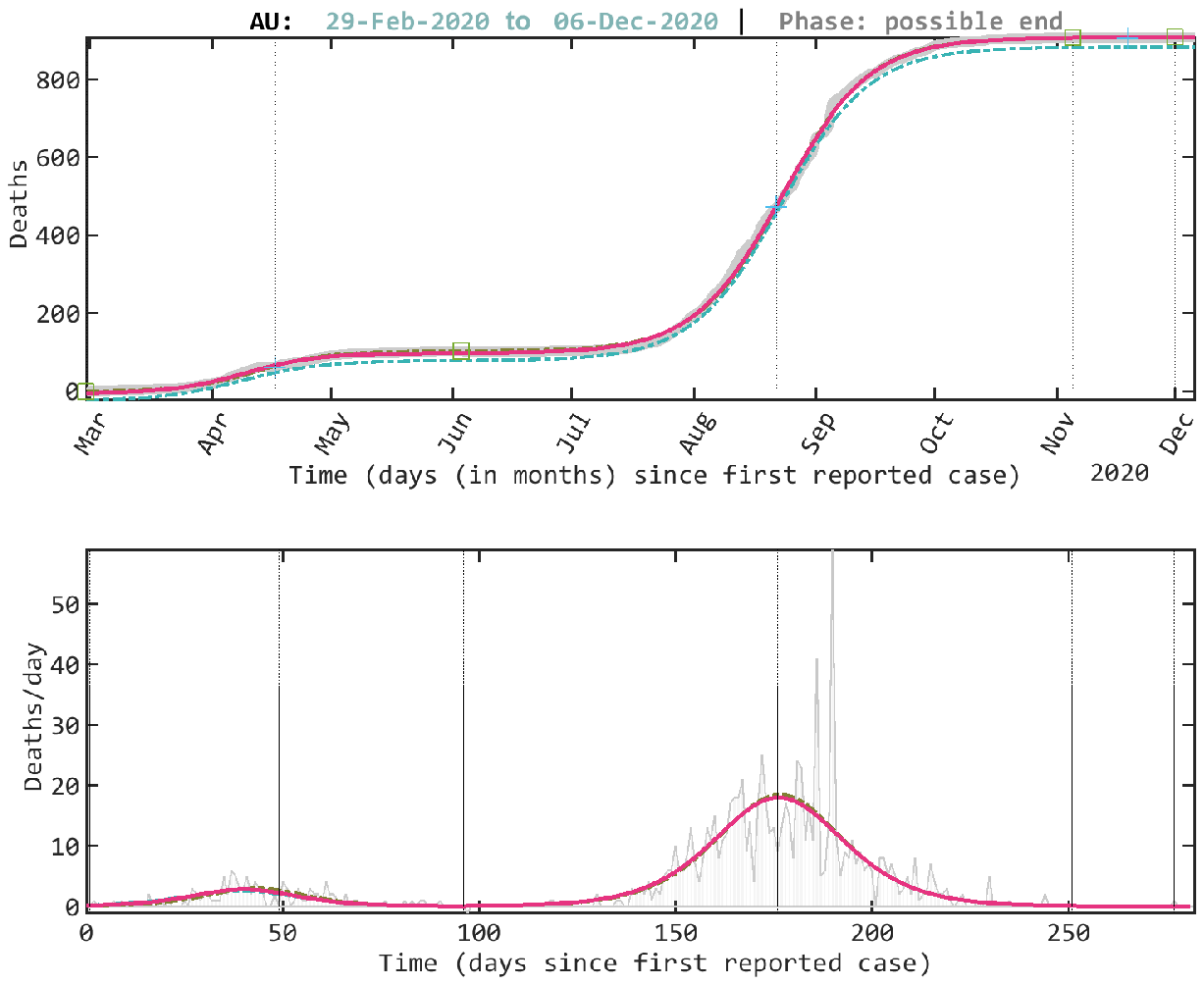}%
		\label{fig:aud}}
	
	\caption{Australia (\ref{fig:aui}) Infections and (\ref{fig:aud}) Deaths: Growth process described by the nlogistic-sigmoid model (red line), actual COVID-19 data (gray lines and areas). Bootstrapped 95\% upper and lower confidence intervals (dashed brown and dashed blue).}
\end{figure}
\begin{figure}[t]
	\centering
	\subfloat[]{\includegraphics[width=0.5\linewidth]{ 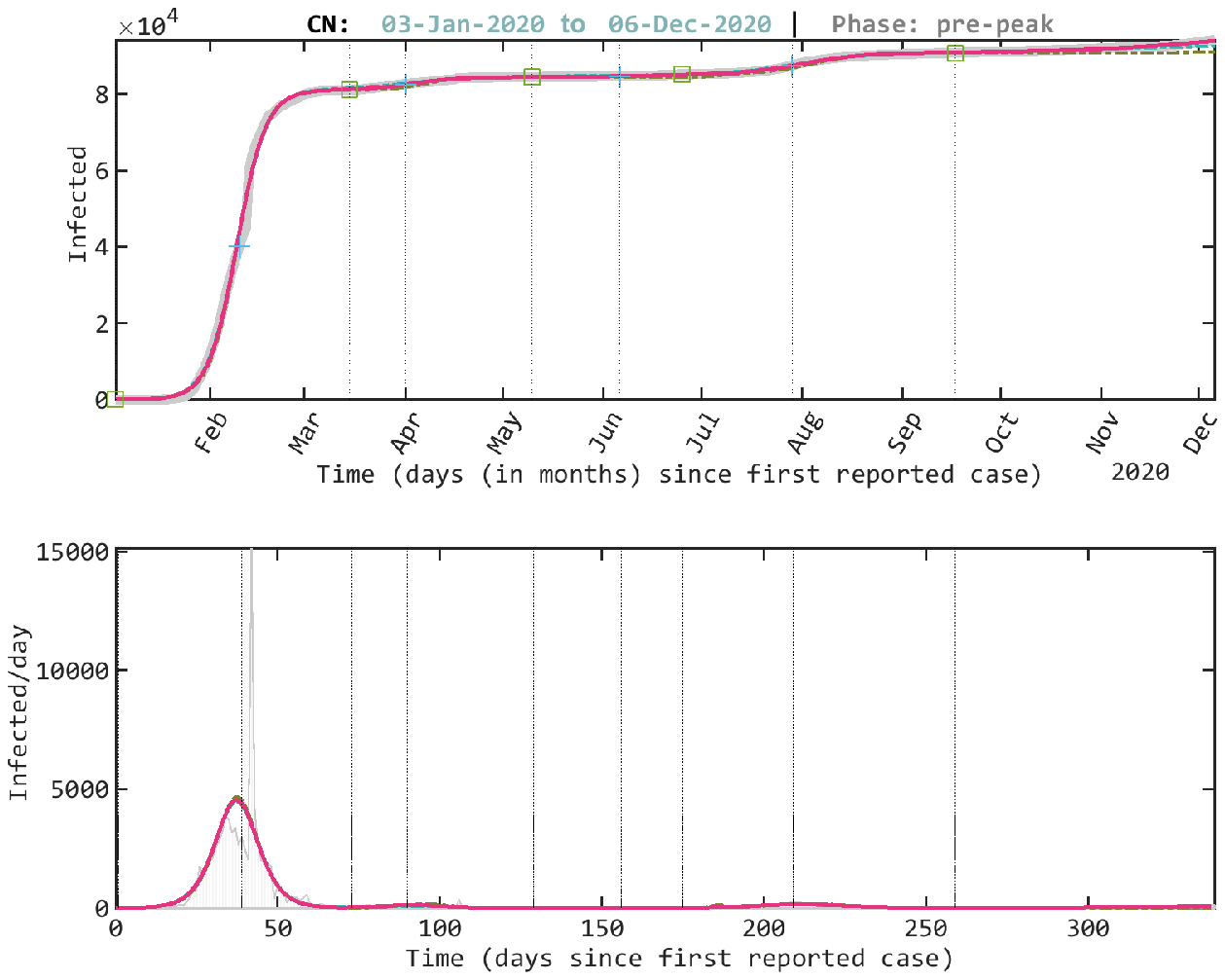}%
		\label{fig:cni}}\quad
	\subfloat[]{\includegraphics[width=0.5\linewidth]{ 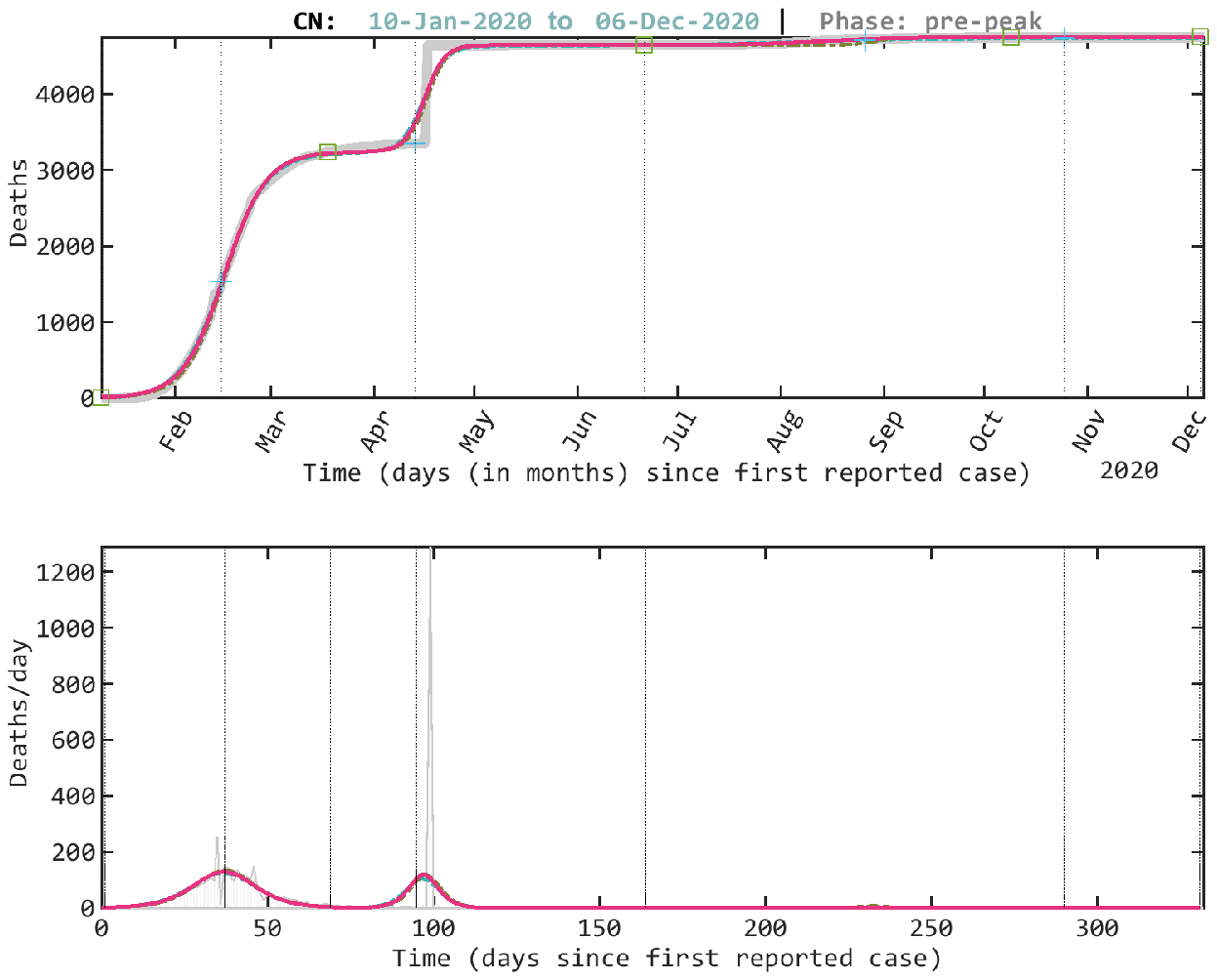}%
		\label{fig:cnd}}
	
	\caption{China (\ref{fig:cni}) Infections and (\ref{fig:cnd}) Deaths: Growth process described by the nlogistic-sigmoid model (red line), actual COVID-19 data (gray lines and areas). Bootstrapped 95\% upper and lower confidence intervals (dashed brown and dashed blue).}
\end{figure}
\begin{figure}[t]
	\centering
	\subfloat[]{\includegraphics[width=0.5\linewidth]{ 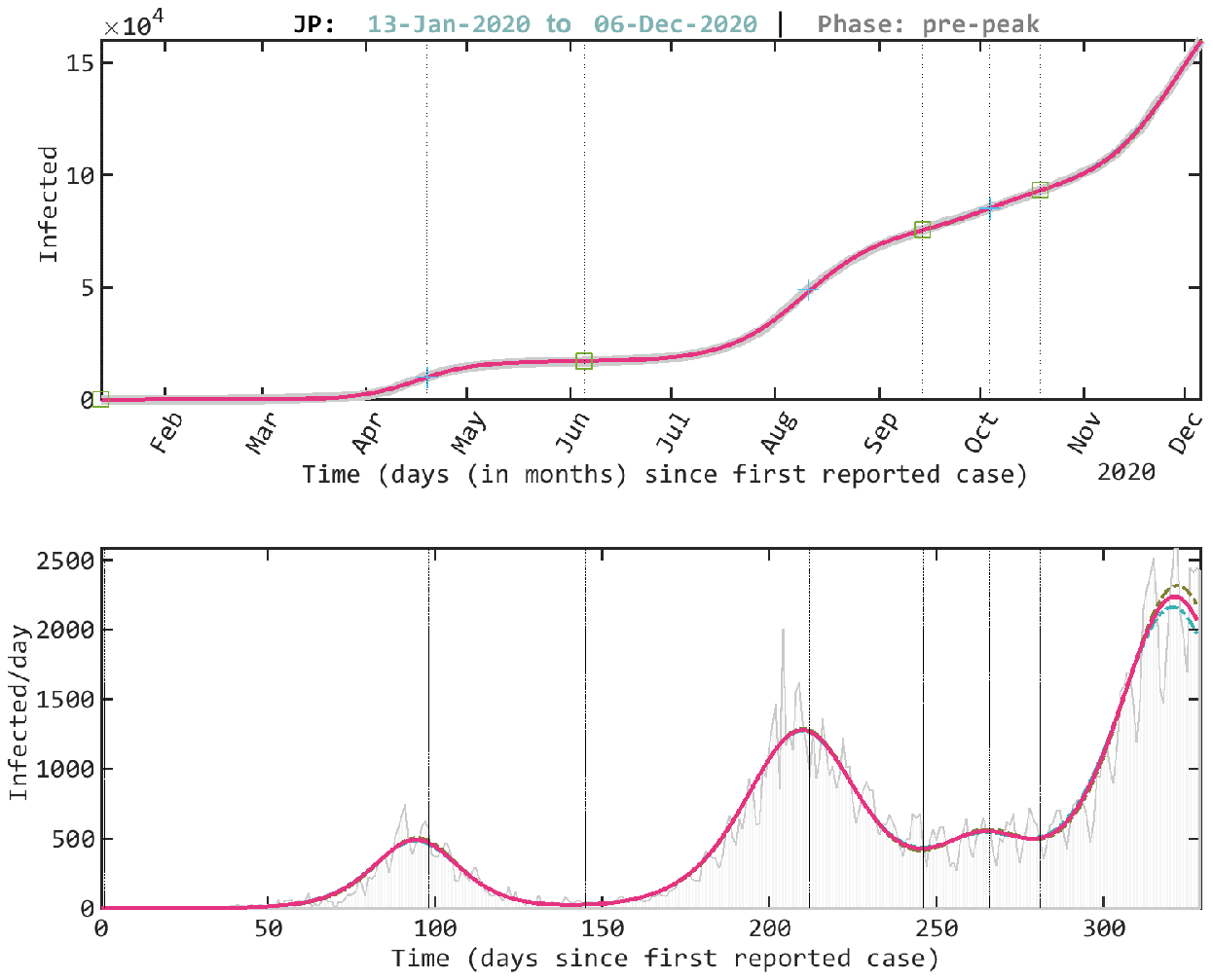}%
		\label{fig:jpi}}\quad
	\subfloat[]{\includegraphics[width=0.5\linewidth]{ 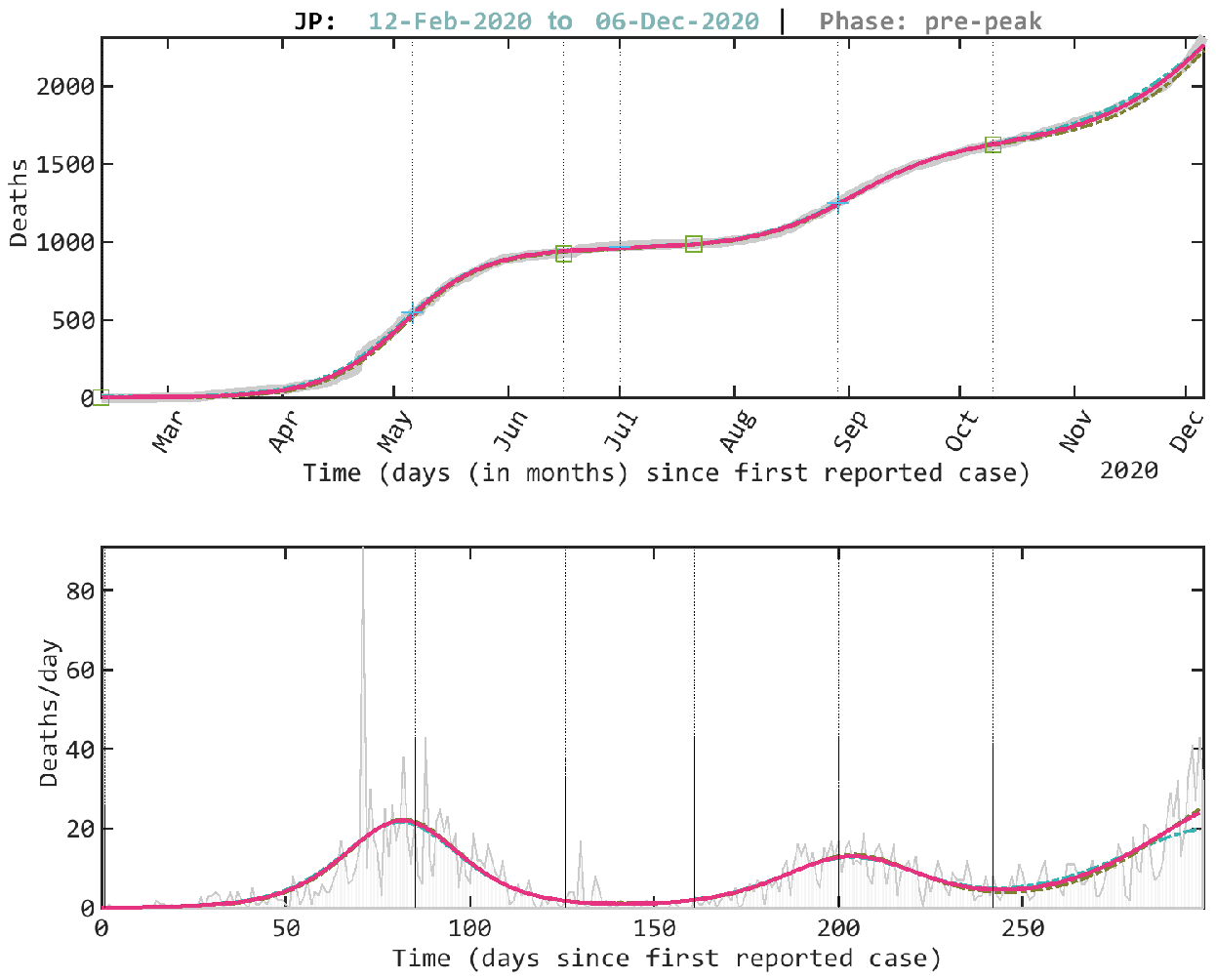}%
		\label{fig:jpd}}
	
	\caption{Japan (\ref{fig:jpi}) Infections and (\ref{fig:jpd}) Deaths: Growth process described by the nlogistic-sigmoid model (red line), actual COVID-19 data (gray lines and areas). Bootstrapped 95\% upper and lower confidence intervals (dashed brown and dashed blue).}
\end{figure}
\begin{figure}[t]
	\centering
	\subfloat[]{\includegraphics[width=0.5\linewidth]{ 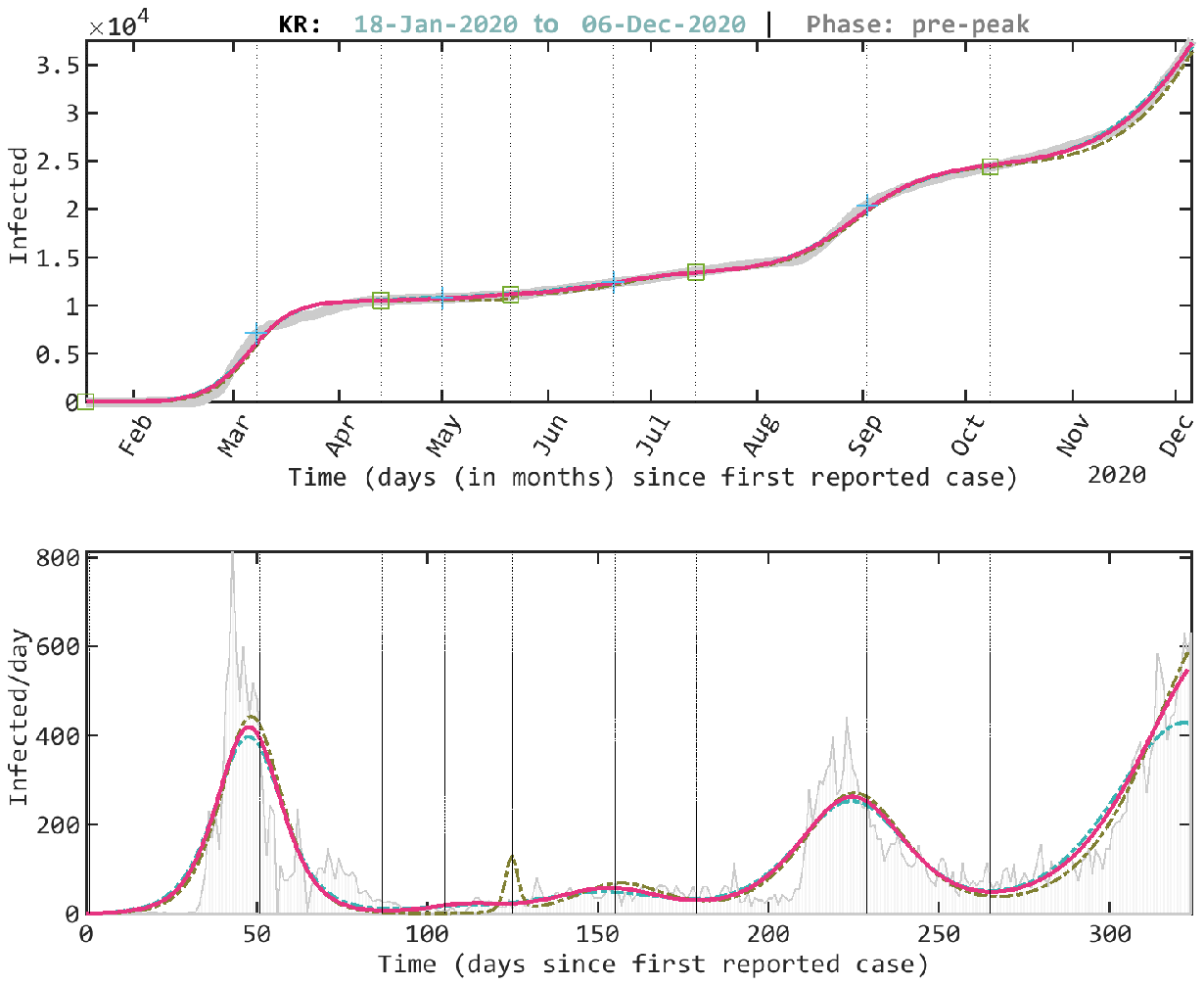}%
		\label{fig:kri}}\quad
	\subfloat[]{\includegraphics[width=0.5\linewidth]{ 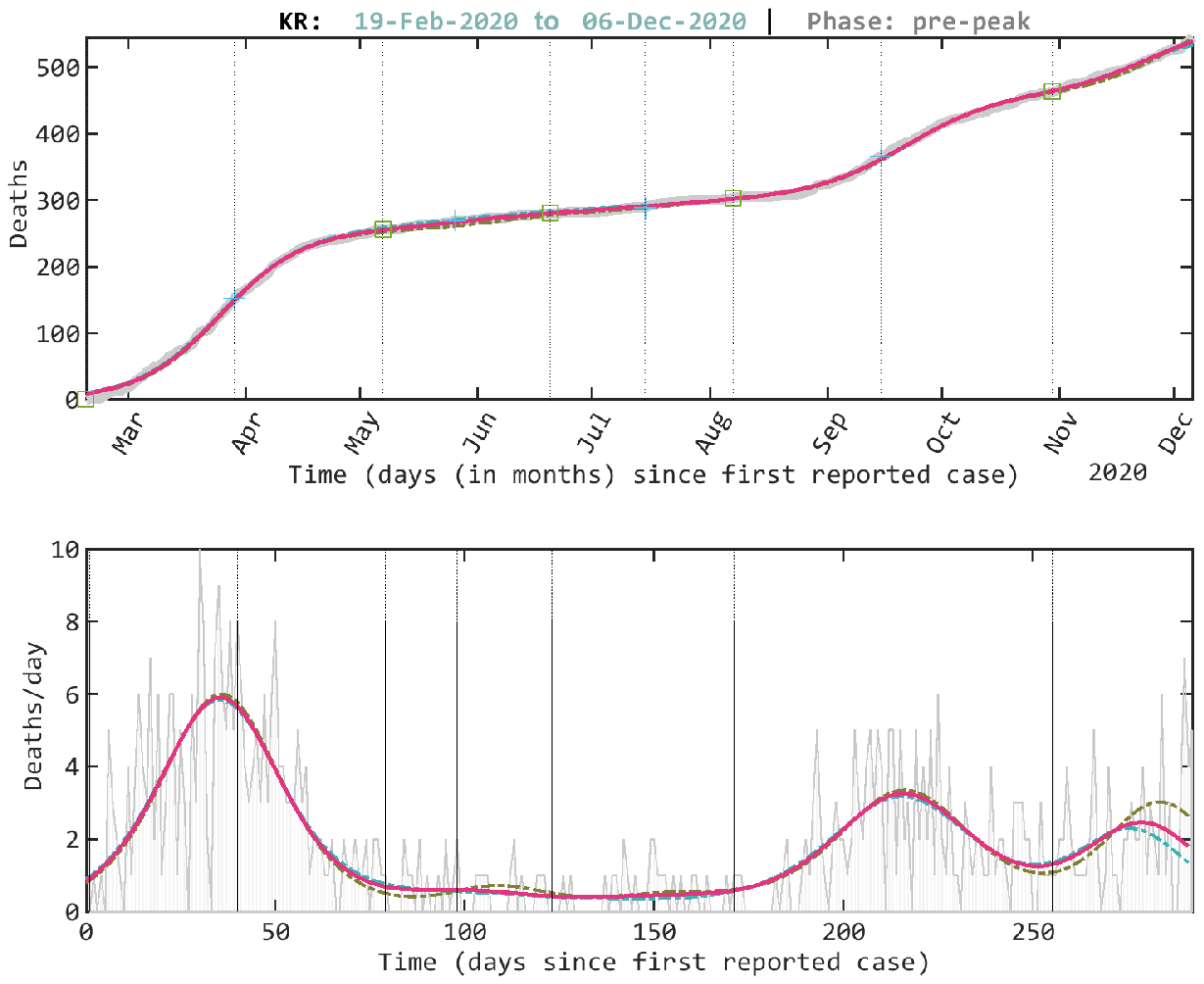}%
		\label{fig:krd}}
	
	\caption{South Korea (\ref{fig:kri}) Infections and (\ref{fig:krd}) Deaths: Growth process described by the nlogistic-sigmoid model (red line), actual COVID-19 data (gray lines and areas). Bootstrapped 95\% upper and lower confidence intervals (dashed brown and dashed blue).}
\end{figure}
\begin{figure}[t]
	\centering
	\subfloat[]{\includegraphics[width=0.5\linewidth]{ 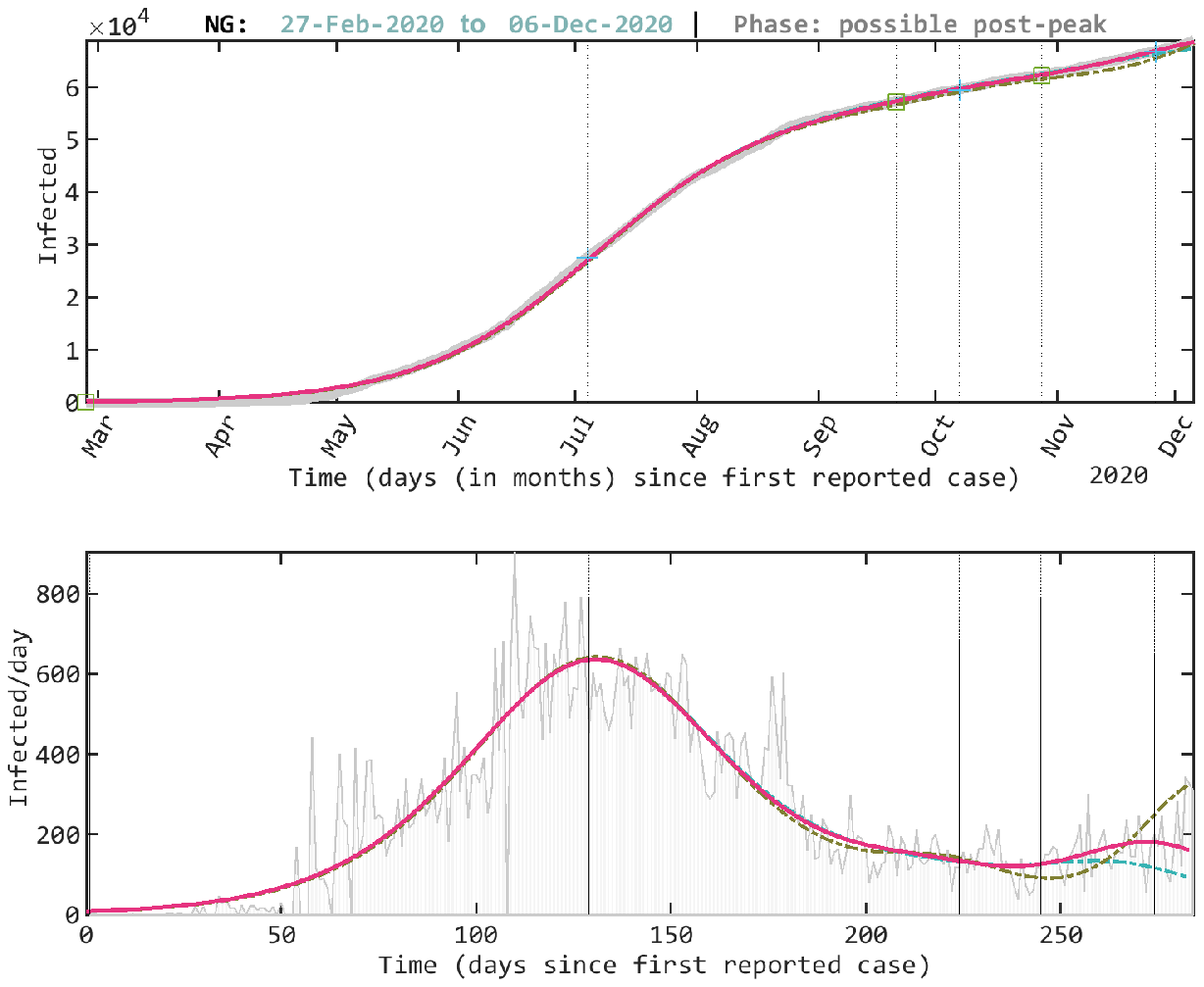}%
		\label{fig:ngi}}\quad
	\subfloat[]{\includegraphics[width=0.5\linewidth]{ 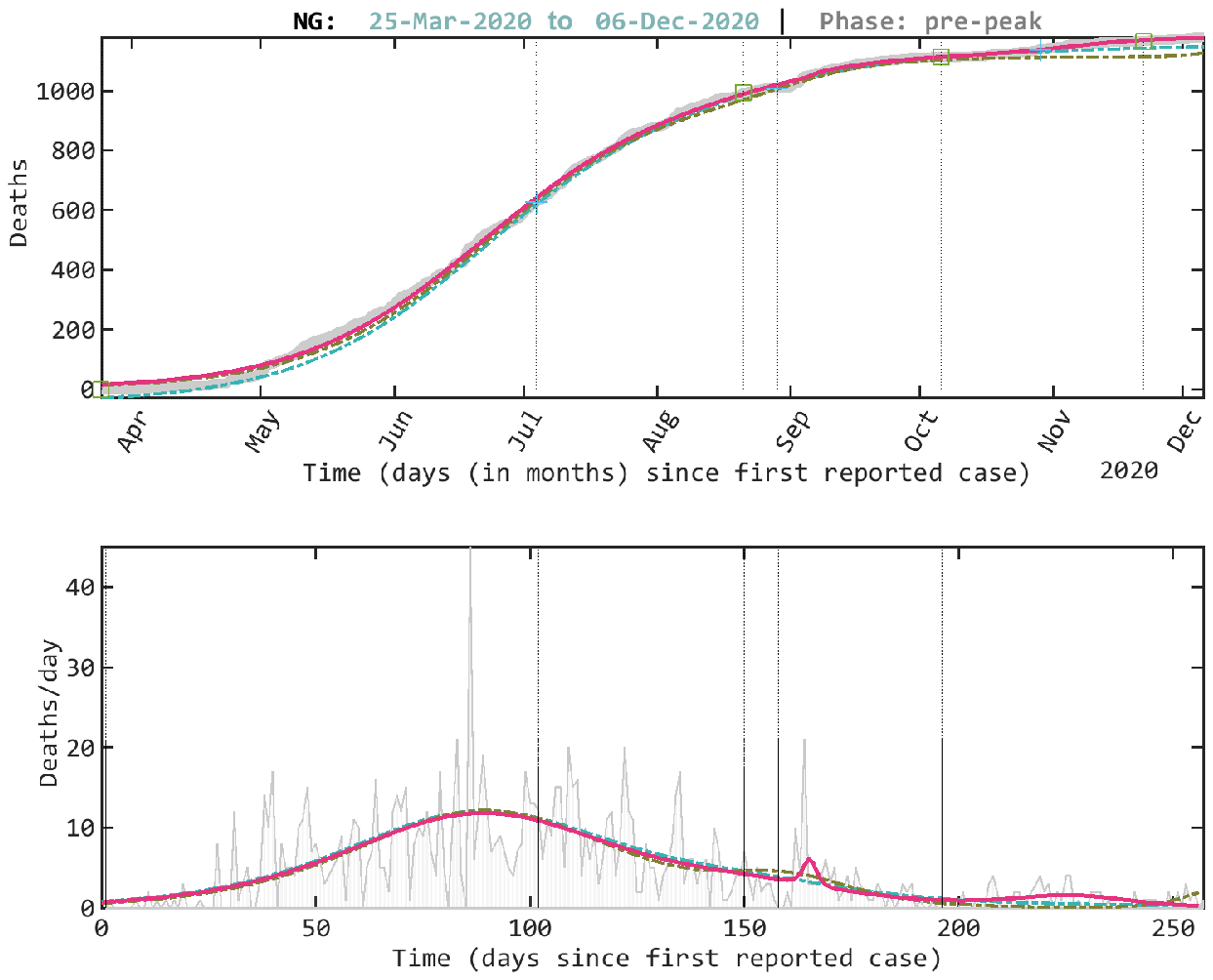}%
		\label{fig:ngd}}
	
	\caption{Nigeria (\ref{fig:ngi}) Infections and (\ref{fig:ngd}) Deaths: Growth process described by the nlogistic-sigmoid model (red line), actual COVID-19 data (gray lines and areas). Bootstrapped 95\% upper and lower confidence intervals (dashed brown and dashed blue).}
\end{figure}
\begin{figure}[t]
	\centering
	\subfloat[]{\includegraphics[width=0.5\linewidth]{ 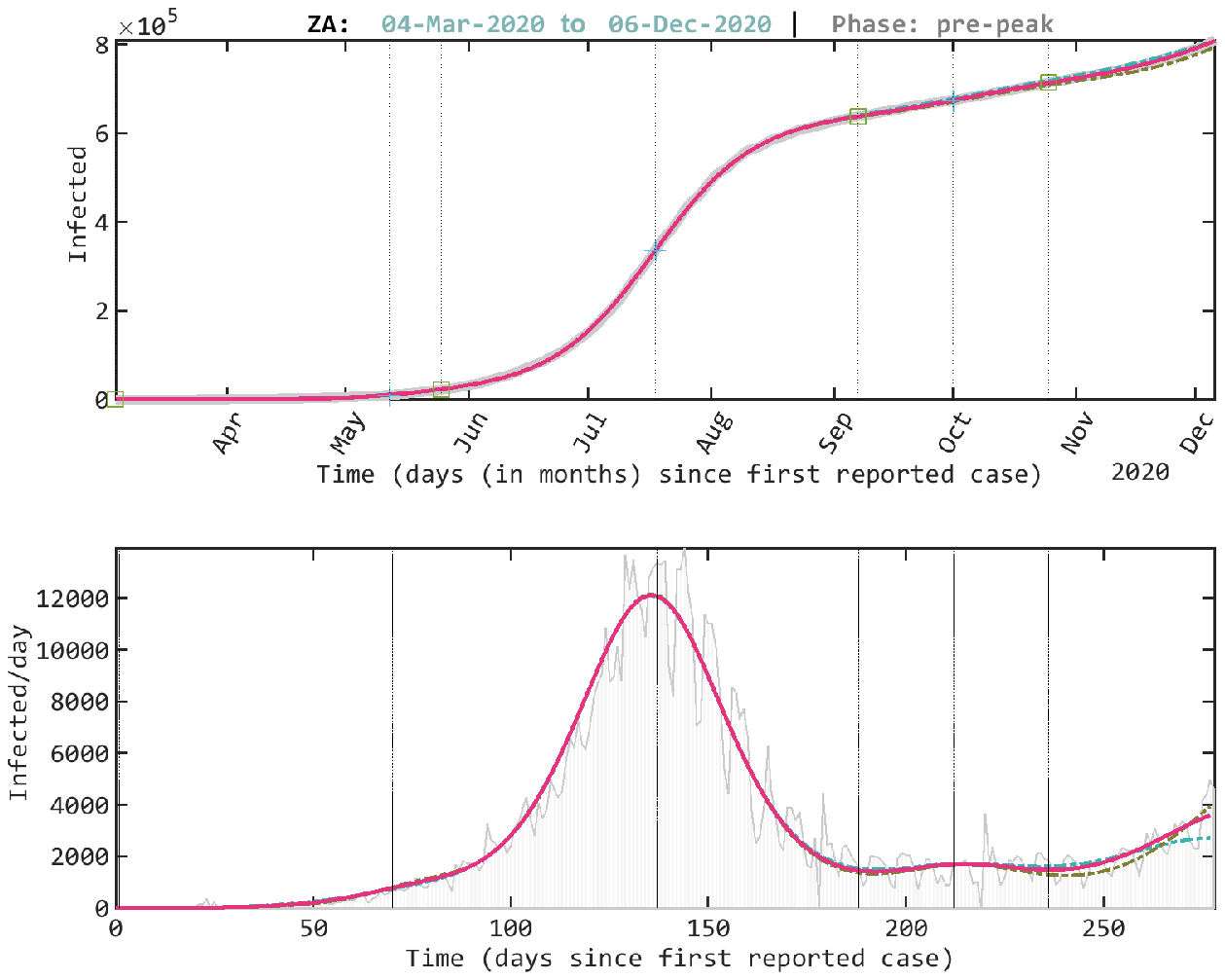}%
		\label{fig:zai}}\quad
	\subfloat[]{\includegraphics[width=0.5\linewidth]{ 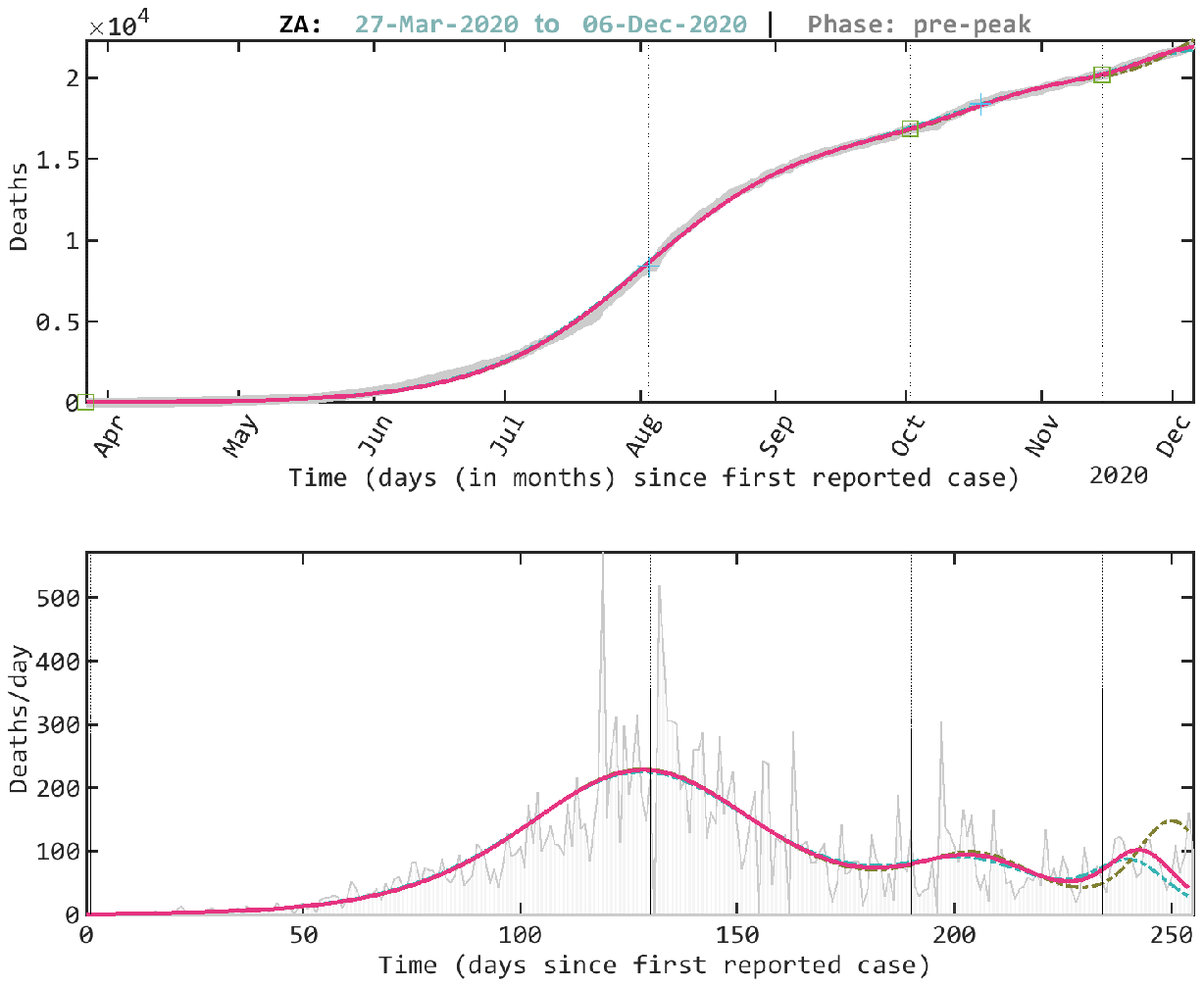}%
		\label{fig:zad}}
	
	\caption{South Africa (\ref{fig:zai}) Infections and (\ref{fig:zad}) Deaths: Growth process described by the nlogistic-sigmoid model (red line), actual COVID-19 data (gray lines and areas). Bootstrapped 95\% upper and lower confidence intervals (dashed brown and dashed blue).}
	\label{fig:za}
\end{figure}

\begin{table*}[t]
	\caption{World: Estimated Logistic-Metrics for the COVID-19 Pandemic}
	\centering
	\def\arraystretch{1.5}%
	\begin{tabular}{|c|c|c|c|c|c|c|}
		\hline
		\multicolumn{4}{|c||}{Infections}&
		\multicolumn{3}{c|}{Deaths}\\
		\hline\hline
			 & $R_2$ & YIR & XIR & $R_2$ & YIR & XIR \\
		\hline\hline
WD&0.9999&	0.4916 [0.4908, 0.5063]& 0.9843 [0.9826, 1.0146]&	
0.9999&	0.4584 [0.4241, 0.5079]& 0.9266 [0.8634, 1.0245]\\	
\hline
\end{tabular}
\label{table_wd}
\end{table*}

\begin{table*}[t]
	\caption{Europe: Estimated Logistic-Metrics for the COVID-19 Pandemic}
	\centering
	\def\arraystretch{1.5}%
	\begin{tabular}{|c|c|c|c|c|c|c|}
	\hline
	\multicolumn{4}{|c||}{Infections}&
	\multicolumn{3}{c|}{Deaths}\\
	\hline\hline
	EU & $R_2$ & YIR & XIR & $R_2$ & YIR & XIR \\
	\hline\hline
	GB& 0.9996& 0.8767 [0.8765, 0.8773]& 2.6744 [2.6728, 2.6762]  &  0.9952 & 0.7127 [0.6960, 0.7616] &  1.5751	[1.5133, 1.7880] \\
	\hline
	IT&  0.9998&  0.9002 [0.8995, 0.9023]& 3.0041 [2.9909, 3.0392]
	  & 0.9976  & 0.6960 [0.6781, 0.7493] &  1.5132	[1.4512, 1.7295] \\
	\hline
	FR&  0.9989&  0.9675 [0.9674, 0.9687]& 5.4632 [5.4586, 5.5719]
	& 0.9982 & 	0.8552	[0.8522, 0.8641] & 2.4307 [2.4008,	2.5253] \\
	\hline
	SE&  0.9995& 0.7671	[0.7669,	0.7717]
	 & 1.8153	[1.8140,	1.8388]& 0.9925& 0.8078	[0.7464, 0.8464]  & 2.0504	[1.7154, 2.3661]\\
	\hline
	TR&  0.9993&  0.3622	[0.3517,	0.4576]
	& 0.7549 [0.7377,	0.9208]& 0.9903	& 0.3641 [0.3523, 0.4590]  & 0.7572 [0.7379, 0.9219] \\
	\hline
	RU&  0.9680& 0.6700	[0.6657,	0.7046]
	 & 1.4263	[1.4122,	1.5465] &0.9993 &  0.5036 [0.4827, 0.5522]&  1.0080 [0.9666, 1.1113] \\
	\hline
\end{tabular}
\label{table_eu}
\end{table*}

\begin{table*}[t]
	\caption{America and Australia: Estimated Logistic-Metrics for the COVID-19 Pandemic}
	\centering
	\def\arraystretch{1.5}%
\begin{tabular}{|c|c|c|c|c|c|c|}
	\hline
	\multicolumn{4}{|c||}{Infections}&
	\multicolumn{3}{c|}{Deaths}\\
	\hline\hline
	AM & $R_2$ & YIR & XIR & $R_2$ & YIR & XIR \\
	\hline\hline
	US&  0.9997		
	&  0.3881 [0.3845, 0.4198]&  0.7965 [0.7905, 0.8508]& 0.9984& 0.5131	[0.2600, 0.6290] & 1.0281	[0.5930, 1.3091] \\
	\hline
	CA&  0.9995		
	&  0.4457 [0.4185, 0.4833]&  0.8969 [0.8484, 0.9675]&  0.9987&  0.4146	[0.3181, 0.5761]& 0.8415 [0.6829, 1.1657]\\
	\hline
	AU&  0.9999	
	& 0.9422 [0.7928, 0.9992]&  2.0478 [0.0126, 262580]&  0.9997		
	& 0.9999 [0.9692, 0.9999]& 0.0013 [0, 0.9985]\\
	\hline
	CU&  0.9992	
	&  0.2515 [0.1497, 0.5890]&  0.5982 [0.4354, 1.3066]&  0.9986 &  0.6770 [0.3908, 0.7333]& 1.4479	[0.8010, 1.6586]  \\
	\hline
	MX&  0.9998	
	&  0.2714 [0.2193, 0.4951]&  0.6574 [0.5713, 1.1138]& 0.9999&  0.4470	[0.2534, 0.5528]& 0.9105 [0.5876, 1.1336] \\
	\hline
	BR& 0.9990		
	 &  0.5161 [0.4622, 0.6624]&  1.1084 [0.9859, 1.5868]& 0.9967		
	  & 0.5140 [0.3800, 0.6092] & 1.1840 [0.8692, 1.5535] \\
	\hline
\end{tabular}
\label{table_amau}
\end{table*}
\begin{table*}[t]
	\caption{Asia: Estimated Logistic-Metrics for the COVID-19 Pandemic}
	\centering
	\def\arraystretch{1.5}%
\begin{tabular}{|c|c|c|c|c|c|c|}
	\hline
	\multicolumn{4}{|c||}{Infections}&
	\multicolumn{3}{c|}{Deaths}\\
	\hline\hline
	AS & $R_2$ & YIR & XIR & $R_2$ & YIR & XIR \\
	\hline\hline
	CN&  0.9976&  0.4464 [0.0092,	0.7745]
	&  0.8981	[0.0965,	1.8637]& 0.9968	
	 & 0.8553	[0.0203, 0.8679] &  0.9983	[0.1440, 2.4367] \\
	\hline
	JP&  0.9999&  0.6373 [0.6234,	0.6488]&  1.3264 [1.2873,	1.3601]	& 0.9994 & 	0.3246[	0.3047,	0.4338] & 0.6950 [0.6625, 0.8797] \\
	\hline
	KR&  0.9981& 	0.3525	[0.3274,	0.5183]
	&  0.7378	[0.6977, 1.0374 ]&  0.9991
	&  	0.7499	[0.6728, 0.8087]&  1.7496 [1.4429, 2.0983]\\
	\hline
	IN&  0.9999& 	0.2607	[0.2198,	0.4138]
	&  0.6875	[0.6225, 1.0211 ]&  0.9996
	& 0.5671 [0.5158, 0.6079] & 1.9248	[1.7914,	1.9400]	 \\
	\hline
	IL&  0.9998&	0.2822	[0.2376,	0.5238]
	 &  0.6287	[0.5598,	1.0537] &  0.9999 & 0.1238	[0.0415, 0.4884]  &  0.4235	[0.2711, 1.3035]\\
	\hline
	IR&  0.9994&  0.4917	[0.4902,	0.5075]& 0.9837	[0.9807,	1.0153]
	 & 0.9999& 0.6765	[0.6690, 0.6834] & 	1.4472	[1.4229, 1.4706] \\
	\hline
	AE&  0.9994&  	0.3929	[0.2141,	0.4668]&  1.0005 [0.6274,	1.2715]
	&  0.9981 &  0.7802	[0.6468, 0.8093]& 2.1061	[1.4185, 2.5238]	 \\
	\hline
\end{tabular}
\label{table_asia}
\end{table*}
\begin{table*}[t]
	\caption{Africa: Estimated Logistic-Metrics for the COVID-19 Pandemic}
	\centering
	\def\arraystretch{1.5}%
\begin{tabular}{|c|c|c|c|c|c|c|}
	\hline
	\multicolumn{4}{|c||}{Infections}&
	\multicolumn{3}{c|}{Deaths}\\
	\hline\hline
	AF & $R_2$ & YIR & XIR & $R_2$ & YIR & XIR \\
	\hline\hline
	NG&  0.9995	
	&  	0.6360 [0.3956, 0.7209]&  1.3615 [0.8200, 1.7139]&  0.9982	&  0.6403 [0.2564, 0.8094]&0.2037 [1.0925E-4, 1.5987E19] \\
	\hline
	GH&  0.9975	
	&  	0.8409 [0.8209, 0.8430]&  2.7727 [2.5154, 2.8286]&  0.9994	&  0.9911	[0.9887, 0.9914]&  10.5468	[9.3726, 10.7498]	\\
	\hline
	EG&  0.9999		
	&  0.2512 [0.1899, 0.4728]&  0.6355 [0.5346,	1.0691]& 0.9999 &  0.7532 [0.7204, 0.7931]& 	1.7544 [1.6095,	1.9720	] \\
	\hline
	ZA&  1.0000	
	&  0.3460 [0.2147, 0.4679]&  0.7316 [0.5254, 0.9457]& 0.9995  & 	0.8586	[0.6476, 0.8763] &  2.7518	[1.3948, 3.1850 ]\\
	\hline
	ZW&  0.9988	
	&  0.3227 [0.2748, 0.5860]&  0.6905 [0.6158, 1.1909]&  0.9986	&  0.7436 [0.5264, 0.7930]& 1.7034	[1.0542,	1.9718]	 \\
	\hline
	KE&  0.9999	
	&  	0.8555 [0.8500,	0.8585]&  2.4334 [2.3811, 2.4636]& 0.9997	 &  0.8322	[0.8296, 0.8360]& 2.2275	[2.2070,	2.2587] \\
	\hline
\end{tabular}
\label{table_africa}
\end{table*}

\subsection{Results}\label{sec_results}
The computed probability-threshold value on all model fits were equivalent to 0, with goodness of fit as high as $99\%$. Therefore we can within a 95\% confidence level accept that the metrics of the \textsc{nlsig} model solution to the COVID-19 data-sets are significant.

Further from \Crefrange{fig:wd}{fig:za}, it is evident that in the selected countries, known to be affected with the COVID-19 epidemic, China has effectively restricted its COVID-19 epidemic growth, compared to most other countries of the world, especially the USA. It is also inferable that the lock-downs and social-distancing mechanisms, imposed by the various governments in these countries, has turned out not to be effective or worth the economic pain, since to date, the COVID-19 growth curve has not saturated (flattened), so as to allow full re-opening of the economy.
For instance, as at this time, some countries have entered recession such as Nigeria, South-Korea, the UK and many others. It will be interesting to observe, in the coming months, whether with the advent of vaccines, the world may finally have some respite.

Below we provide textual interpretation for the bootstrapped 95\% confidence intervals logistic-metrics, illustrated in \Crefrange{table_wd}{table_africa} for the COVID-19 pandemic as at 6th December, 2020. It should be noted that these estimations are based on the optimization solution at the current time, and may change as more data is received. Also the metrics can have a wide array of intersecting meanings when the confidence intervals fall in a different range. Notably, these metrics do not indicate by what amount the growth is increasing or decreasing, or whether an end has occurred.

\textbf{World}: For infections: the YIR indicates that the numbers are peaking and may start to decrease soon; the XIR indicates that this time is most likely a peak period. For deaths: the YIR indicates that the numbers are still increasing but may likely peak soon; the XIR indicates that this time may most likely be a peak period but could also be a pre-peak period.

\textbf{United Kingdom and Italy}: For both infections and deaths: the YIR indicates that the numbers are now decreasing; the XIR indicates that this time is clearly a post-peak period.

\textbf{Nigeria}: For infections: the YIR indicates that the numbers are decreasing compared to the past but may most likely also be increasing; the XIR indicates that at this time, the growth is in a post-peak period but could again most likely be a pre-peak period. For the deaths: the YIR indicates that the numbers are most likely decreasing compared to the past but could possibly have started increasing; the XIR indicates that the growth at this time is clearly in a post-peak period.

\textbf{South-Africa}: For infections: the YIR indicates that the numbers are still increasing; the XIR indicates that this time is a pre-peak period. For the deaths: the YIR indicates that the numbers are decreasing; the XIR indicates that this time is a post-peak period.

\textbf{United States of America}: For infections: the YIR indicates that the numbers are still increasing; the XIR indicates that this time is a pre-peak period. For the deaths: the YIR indicates that the numbers could most likely be decreasing but may also, still be increasing; the XIR indicates that the growth has entered a post-peak period, but may still be in a pre-peak period.

\textbf{Canada}: For both infections and deaths: the YIR indicates that the numbers are still increasing but may peak soon; the XIR indicates that at this time, the growth may soon enter its peak period.

\textbf{Australia}: For both infections and deaths: the YIR indicates that the numbers are now decreasing; the XIR indicates that this time is clearly a post-peak period.

\textbf{China}: For infections and deaths: the YIR indicates that the numbers are generally most-likely reducing; the XIR indicates that this time is most-likely a post-peak period.

\textbf{Japan}: For infections: the YIR indicates that the numbers are now decreasing; the XIR indicates that this time is a post-peak period. For the deaths: the YIR indicates that the numbers are still increasing; the XIR indicates that this time is a pre-peak period.

\textbf{South Korea}: For infections: the YIR indicates that the numbers are increasing or may have reached a peak; the XIR indicates that this time is a pre-peak period close to a peak period. For the deaths: the YIR indicates that the numbers are decreasing; the XIR indicates that this time is a post-peak period.

Generally, it can be observed that the logistic metrics when combined from both perspectives give a more robust interpretation of the growth. Therefore, compared to many models, which engage in the risky and costly acts of prophecy, we propose these metrics as a safer projection, which public-policy makers and other professionals can apply in monitoring the COVID-19 pandemic growth phenomenon (infections or deaths) and the effectiveness of containment and vaccination strategies, till the growth flattens out completely in a particular locale or in the world. Such modelling  setup as proposed in this paper is available at the repository \cite{somefunSomefunAgbaNLSIGCOVID19Lab2020}.

The only currently noted limitation of the logistic-growth metrics lies with the \textsc{nlsig} model solution found by the optimization algorithm or method.

\section{Conclusions}\label{sec_concls}
We have presented a modern definition for the classic logistic function, called the \textbf{n}logistic-sigmoid function that can be used to model the uncertain, noisy, multiple growth behaviour of real-world processes, such as the COVID-19 pandemic, with a high degree of accuracy. This logistic function definition, itself, was shown to be a neural network. Two logistic metrics, the YIR and XIR, were newly introduced for robust monitoring of growth from its two-dimensional perspective. Our modelling results and the estimated logistic metrics indicate that the COVID-19 pandemic growth follows, essentially, the principle of logistic growth. In all applications, the model solutions provided statistically significant goodness of fit, often as high as $99\%$, to the dataset describing the COVID-19 epidemic spread in the world and each of the selected locales. We strongly believe that the \textsc{nlsig} can again increase the value of the logistic-sigmoid as a powerful  machine learning tool, for modelling and regression, in many areas such as biomedical and epidemiological research and other scientific and engineering disciplines.

\bibliographystyle{../bib/naturemag}
\bibliography{../bib/Nlsig}

\begin{thebibliography}{10}
\expandafter\ifx\csname url\endcsname\relax
  \def\url#1{\texttt{#1}}\fi
\expandafter\ifx\csname urlprefix\endcsname\relax\def\urlprefix{URL }\fi
\providecommand{\bibinfo}[2]{#2}
\providecommand{\eprint}[2][]{\url{#2}}

\bibitem{somefunSomefunAgbaNLSIGCOVID19Lab2020}
\bibinfo{author}{Somefun, O.}
\newblock \bibinfo{title}{{{somefunAgba}}/{{NLSIG}}\_{{COVID19Lab}}}
  (\bibinfo{year}{2020}).
\newblock \urlprefix\url{https://github.com/somefunAgba/NLSIG_COVID19Lab}.

\bibitem{shulmanMathAliveUsingOriginal1998}
\bibinfo{author}{Shulman, B.}
\newblock \bibinfo{title}{Math-{{Alive}}! {{Using Original Sources To Teach
  Mathematics In Social Context}}}.
\newblock \emph{\bibinfo{journal}{PRIMUS}} \textbf{\bibinfo{volume}{8}},
  \bibinfo{pages}{1--14} (\bibinfo{year}{1998}).

\bibitem{jefferyRestrictedLogarithmicGrowth2016}
\bibinfo{author}{Jeffery, P.~F.}
\newblock \bibinfo{title}{Restricted {{Logarithmic Growth}} with {{Injection}}}
  (\bibinfo{year}{2016}).
\newblock
  \urlprefix\url{http://JeffreyFreeman.me/restricted-logarithmic-growth-with-injection/}.

\bibitem{udellMaximizingSumSigmoids2013}
\bibinfo{author}{Udell, M.} \& \bibinfo{author}{Boyd, S.}
\newblock \bibinfo{title}{Maximizing a sum of sigmoids}.
\newblock \emph{\bibinfo{journal}{Optimization and Engineering}}
  \bibinfo{pages}{1--25} (\bibinfo{year}{2013}).

\bibitem{cybenkoApproximationSuperpositionsSigmoidal1989}
\bibinfo{author}{Cybenko, G.}
\newblock \bibinfo{title}{Approximation by superpositions of a sigmoidal
  function}.
\newblock \emph{\bibinfo{journal}{Mathematics of Control, Signals and Systems}}
  \textbf{\bibinfo{volume}{2}}, \bibinfo{pages}{303--314}
  (\bibinfo{year}{1989}).

\bibitem{lipovetsky_double_2010}
\bibinfo{author}{Lipovetsky, S.}
\newblock \bibinfo{title}{Double logistic curve in regression modeling}.
\newblock \emph{\bibinfo{journal}{Journal of Applied Statistics}}
  \textbf{\bibinfo{volume}{37}}, \bibinfo{pages}{1785--1793}
  (\bibinfo{year}{2010}).

\bibitem{szeEfficientProcessingDeep2017}
\bibinfo{author}{Sze, V.}, \bibinfo{author}{Chen, Y.}, \bibinfo{author}{Yang,
  T.} \& \bibinfo{author}{Emer, J.~S.}
\newblock \bibinfo{title}{Efficient {{Processing}} of {{Deep Neural Networks}}:
  {{A Tutorial}} and {{Survey}}}.
\newblock \emph{\bibinfo{journal}{Proceedings of the IEEE}}
  \textbf{\bibinfo{volume}{105}}, \bibinfo{pages}{2295--2329}
  (\bibinfo{year}{2017}).

\bibitem{barkerLogisticFunctionComplementary2020}
\bibinfo{author}{Barker, J.} \& \bibinfo{author}{Watt, J.}
\newblock \bibinfo{title}{Logistic function and complementary {{Fermi}}
  function for simple modelling of the {{COVID}}-19 pandemic: Tutorial notes}
  (\bibinfo{year}{2020}).

\bibitem{christopoulosEfficientIdentificationInflection2016}
\bibinfo{author}{Christopoulos, D.~T.}
\newblock \bibinfo{title}{On the efficient identification of an inflection
  point}.
\newblock \emph{\bibinfo{journal}{International Journal of Mathematics and
  Scientific Computing,(ISSN: 2231-5330)}} \textbf{\bibinfo{volume}{6}}
  (\bibinfo{year}{2016}).

\bibitem{yeoModellingTechniqueUtilizing2016}
\bibinfo{author}{Yeo, H.~L.} \& \bibinfo{author}{Tseng, K.~J.}
\newblock \bibinfo{title}{Modelling technique utilizing modified sigmoid
  functions for describing power transistor device capacitances applied on
  {{GaN HEMT}} and silicon {{MOSFET}}}.
\newblock In \emph{\bibinfo{booktitle}{2016 {{IEEE Applied Power Electronics
  Conference}} and {{Exposition}} ({{APEC}})}}, \bibinfo{pages}{3107--3114}
  (\bibinfo{year}{2016}).

\bibitem{hemmatiFairEfficientBandwidth2016}
\bibinfo{author}{Hemmati, M.}, \bibinfo{author}{McCormick, B.} \&
  \bibinfo{author}{Shirmohammadi, S.}
\newblock \bibinfo{title}{Fair and efficient bandwidth allocation for video
  flows using sigmoidal programming}.
\newblock In \emph{\bibinfo{booktitle}{2016 {{IEEE International Symposium}} on
  {{Multimedia}} ({{ISM}})}}, \bibinfo{pages}{226--231}
  (\bibinfo{publisher}{{IEEE}}, \bibinfo{year}{2016}).

\bibitem{xsRichardsModelRevisited2012}
\bibinfo{author}{Xs, W.}, \bibinfo{author}{J, W.} \& \bibinfo{author}{Y, Y.}
\newblock \bibinfo{title}{Richards model revisited: Validation by and
  application to infection dynamics}.
\newblock \emph{\bibinfo{journal}{Journal of theoretical biology}}
  \textbf{\bibinfo{volume}{313}} (\bibinfo{year}{2012}).

\bibitem{shaoNonlinearTrackingDifferentiator20140901}
\bibinfo{author}{Shao, X.} \& \bibinfo{author}{Wang, H.}
\newblock \bibinfo{title}{{Nonlinear tracking differentiator based on improved
  sigmoid function}}.
\newblock \emph{\bibinfo{journal}{Control Theory and Application}}
  \textbf{\bibinfo{volume}{31}}, \bibinfo{pages}{1116--1122}
  (\bibinfo{year}{20140901}).

\bibitem{yinFlexibleSigmoidFunction2003}
\bibinfo{author}{Yin, X.}, \bibinfo{author}{Goudriaan, J.},
  \bibinfo{author}{Lantinga, E.~A.}, \bibinfo{author}{Vos, J.} \&
  \bibinfo{author}{Spiertz, H.~J.}
\newblock \bibinfo{title}{A {{Flexible Sigmoid Function}} of {{Determinate
  Growth}}}.
\newblock \emph{\bibinfo{journal}{Annals of Botany}}
  \textbf{\bibinfo{volume}{91}}, \bibinfo{pages}{361--371}
  (\bibinfo{year}{2003}).

\bibitem{leeEstimationCOVID19Spread2020}
\bibinfo{author}{Lee, S.~Y.}, \bibinfo{author}{Lei, B.} \&
  \bibinfo{author}{Mallick, B.}
\newblock \bibinfo{title}{Estimation of {{COVID}}-19 spread curves integrating
  global data and borrowing information}.
\newblock \emph{\bibinfo{journal}{PLOS ONE}} \textbf{\bibinfo{volume}{15}},
  \bibinfo{pages}{e0236860} (\bibinfo{year}{2020}).

\bibitem{wuGeneralizedLogisticGrowth2020}
\bibinfo{author}{Wu, K.}, \bibinfo{author}{Darcet, D.}, \bibinfo{author}{Wang,
  Q.} \& \bibinfo{author}{Sornette, D.}
\newblock \bibinfo{title}{Generalized logistic growth modeling of the
  {{COVID}}-19 outbreak: Comparing the dynamics in the 29 provinces in
  {{China}} and in the rest of the world}.
\newblock \emph{\bibinfo{journal}{Nonlinear Dynamics}}
  \textbf{\bibinfo{volume}{101}}, \bibinfo{pages}{1561--1581}
  (\bibinfo{year}{2020}).

\bibitem{loibelInferenceRichardsGrowth2006}
\bibinfo{author}{Loibel, S.}, \bibinfo{author}{{do Val}, J. B.~R.} \&
  \bibinfo{author}{Andrade, M.~G.}
\newblock \bibinfo{title}{Inference for the {{Richards}} growth model using
  {{Box}} and {{Cox}} transformation and bootstrap techniques}.
\newblock \emph{\bibinfo{journal}{Ecological Modelling}}
  \textbf{\bibinfo{volume}{191}}, \bibinfo{pages}{501--512}
  (\bibinfo{year}{2006}).

\bibitem{tabatabaiHyperbolasticGrowthModels2005}
\bibinfo{author}{Tabatabai, M.}, \bibinfo{author}{Williams, D.~K.} \&
  \bibinfo{author}{Bursac, Z.}
\newblock \bibinfo{title}{Hyperbolastic growth models: Theory and application}.
\newblock \emph{\bibinfo{journal}{Theoretical Biology and Medical Modelling}}
  \textbf{\bibinfo{volume}{2}}, \bibinfo{pages}{14} (\bibinfo{year}{2005}).

\bibitem{taylorForecastingScale2018}
\bibinfo{author}{Taylor, S.~J.} \& \bibinfo{author}{Letham, B.}
\newblock \bibinfo{title}{Forecasting at {{Scale}}}.
\newblock \emph{\bibinfo{journal}{The American Statistician}}
  \textbf{\bibinfo{volume}{72}}, \bibinfo{pages}{37--45}
  (\bibinfo{year}{2018}).

\bibitem{bejanConstructalLawOrigin2011}
\bibinfo{author}{Bejan, A.} \& \bibinfo{author}{Lorente, S.}
\newblock \bibinfo{title}{The constructal law origin of the logistics {{S}}
  curve}.
\newblock \emph{\bibinfo{journal}{Journal of Applied Physics}}
  \textbf{\bibinfo{volume}{110}}, \bibinfo{pages}{024901}
  (\bibinfo{year}{2011}).

\bibitem{reedSummationLogisticCurves1927}
\bibinfo{author}{Reed, L.~J.} \& \bibinfo{author}{Pearl, R.}
\newblock \bibinfo{title}{On the {{Summation}} of {{Logistic Curves}}}.
\newblock \emph{\bibinfo{journal}{Journal of the Royal Statistical Society}}
  \textbf{\bibinfo{volume}{90}}, \bibinfo{pages}{729--746}
  (\bibinfo{year}{1927}).

\bibitem{hanInfluenceSigmoidFunction1995}
\bibinfo{author}{Han, J.} \& \bibinfo{author}{Moraga, C.}
\newblock \bibinfo{title}{The influence of the sigmoid function parameters on
  the speed of backpropagation learning}.
\newblock In \bibinfo{editor}{Mira, J.} \& \bibinfo{editor}{Sandoval, F.}
  (eds.) \emph{\bibinfo{booktitle}{From {{Natural}} to {{Artificial Neural
  Computation}}}}, Lecture {{Notes}} in {{Computer Science}},
  \bibinfo{pages}{195--201} (\bibinfo{publisher}{{Springer}},
  \bibinfo{address}{{Berlin, Heidelberg}}, \bibinfo{year}{1995}).

\bibitem{batistaEstimationStateCorona2020}
\bibinfo{author}{Batista, M.}
\newblock \bibinfo{title}{Estimation of a state of {{Corona}} 19 epidemic in
  {{August}} 2020 by multistage logistic model: A case of {{EU}}, {{USA}}, and
  {{World}} ({{Update September}} 2020)}.
\newblock \emph{\bibinfo{journal}{medRxiv}}
  \bibinfo{pages}{2020.08.31.20185165} (\bibinfo{year}{2020}).

\bibitem{hsiehRealtimeForecastMultiphase2006}
\bibinfo{author}{Hsieh, Y.-H.} \& \bibinfo{author}{Cheng, Y.-S.}
\newblock \bibinfo{title}{Real-time {{Forecast}} of {{Multiphase Outbreak}}}.
\newblock \emph{\bibinfo{journal}{Emerging Infectious Diseases}}
  \textbf{\bibinfo{volume}{12}}, \bibinfo{pages}{122--127}
  (\bibinfo{year}{2006}).

\bibitem{chowellNovelSubepidemicModeling2019}
\bibinfo{author}{Chowell, G.}, \bibinfo{author}{Tariq, A.} \&
  \bibinfo{author}{Hyman, J.~M.}
\newblock \bibinfo{title}{A novel sub-epidemic modeling framework for
  short-term forecasting epidemic waves}.
\newblock \emph{\bibinfo{journal}{BMC Medicine}} \textbf{\bibinfo{volume}{17}},
  \bibinfo{pages}{164} (\bibinfo{year}{2019}).

\bibitem{werbosBackpropagationTimeWhat1990}
\bibinfo{author}{Werbos, P.}
\newblock \bibinfo{title}{Backpropagation through time: What it does and how to
  do it}.
\newblock \emph{\bibinfo{journal}{Proceedings of the IEEE}}
  \textbf{\bibinfo{volume}{78}}, \bibinfo{pages}{1550--1560}
  (\bibinfo{year}{1990}).

\bibitem{okabeMathematicalModelEpidemics2020}
\bibinfo{author}{Okabe, Y.} \& \bibinfo{author}{Shudo, A.}
\newblock \bibinfo{title}{A {{Mathematical Model}} of
  {{Epidemics}}\textemdash{{A Tutorial}} for {{Students}}}.
\newblock \emph{\bibinfo{journal}{Mathematics}} \textbf{\bibinfo{volume}{8}},
  \bibinfo{pages}{1174} (\bibinfo{year}{2020}).

\bibitem{christopoulosNovelApproachEstimating2020}
\bibinfo{author}{Christopoulos, D.~T.}
\newblock \bibinfo{title}{A {{Novel Approach}} for {{Estimating}} the {{Final
  Outcome}} of {{Global Diseases Like COVID}}-19}.
\newblock \emph{\bibinfo{journal}{medRxiv}}
  \bibinfo{pages}{2020.07.03.20145672} (\bibinfo{year}{2020}).

\bibitem{matthewWhyModelingSpread2020}
\bibinfo{author}{Matthew, H.}
\newblock \bibinfo{title}{Why {{Modeling}} the {{Spread}} of {{COVID}}-19 {{Is
  So Damn Hard}} - {{IEEE Spectrum}}} (\bibinfo{year}{2020}).
\newblock
  \urlprefix\url{https://spectrum.ieee.org/artificial-intelligence/medical-ai/why-modeling-the-spread-of-covid19-is-so-damn-hard}.

\end{thebibliography}

\end{document}